\documentclass[lettersize,journal]{IEEEtran}
\usepackage{amsmath,amsfonts}
\usepackage{algorithmic}
\usepackage{algorithm}
\usepackage{array}
\usepackage[caption=false,font=normalsize,labelfont=sf,textfont=sf]{subfig}
\usepackage{textcomp}
\usepackage{stfloats}
\usepackage{url}
\usepackage{verbatim}
\usepackage{graphicx}
\usepackage{cite}
\usepackage{graphicx}
\usepackage{pifont}
\newcommand{\cmark}{\ding{51}}%
\newcommand{\xmark}{\ding{55}}%
\usepackage{algorithmic}
\usepackage{algorithm}
\usepackage{wrapfig,lipsum,booktabs}
\usepackage{amsmath,amssymb,amsfonts}
\usepackage{amsthm}
\usepackage{bbm}
\usepackage{hyperref}
\usepackage{url}
\newtheorem{definition}{Definition}
\newtheorem{theorem}{Theorem}

\newtheorem{remark}{Remark}

\usepackage{multirow}
\usepackage[inline]{enumitem}   
\usepackage{color}
\usepackage{booktabs} 
\usepackage{subcaption} 
\newcommand{\revision}[1]{{\color{black} #1}}
\newcommand{\revisionfinal}[1]{{\color{black} #1}}

\ifCLASSINFOpdf
\else
\fi

\begin{document}
%
\title{\textit{SPLITZ}: Certifiable Robustness via Split Lipschitz Randomized Smoothing\\

\thanks{This work was supported by NSF grants CAREER 1651492, CCF-2100013, CNS-2209951, CNS-1822071, CNS-2317192, and by the U.S. Department of Energy, Office of Science, Office of Advanced Scientific Computing under Award Number DE-SC-ERKJ422, and NIH Award R01-CA261457-01A1.}}

%
%
\author{\IEEEauthorblockA{Meiyu Zhong ~~ Ravi Tandon}

\IEEEauthorblockA{Department of Electrical and Computer Engineering \\
University of Arizona, Tucson, USA \\
E-mail: \textit{\{meiyuzhong, tandonr\}}@arizona.edu
}}


\maketitle
 
\begin{abstract}
Certifiable robustness gives the guarantee that small perturbations around an input to a classifier will not change the prediction. There are two approaches to provide certifiable robustness to adversarial examples-- a) explicitly training classifiers with small Lipschitz constants, and b) Randomized smoothing, which adds random noise to the input to create a smooth classifier. We propose \textit{SPLITZ}, a practical and novel approach which leverages the synergistic benefits of both the above ideas into a single framework. Our main idea is to \textit{split} a classifier into two halves, constrain the Lipschitz constant of the first half, and smooth the second half via randomization. Motivation for \textit{SPLITZ} comes from the observation that many standard deep networks exhibit heterogeneity in Lipschitz constants across layers. \textit{SPLITZ} can exploit this heterogeneity while inheriting the scalability of randomized smoothing. 
We present a principled approach to train \textit{SPLITZ} and provide theoretical analysis to derive certified robustness guarantees during inference. 
\revision{We present a comprehensive comparison of robustness-accuracy trade-offs and show that \textit{SPLITZ} consistently improves on existing state-of-the-art approaches in the MNIST, CIFAR-10 and ImageNet datasets. For instance, with $\ell_2$ norm perturbation budget of \textbf{$\epsilon=1$}, \textit{SPLITZ} achieves $\textbf{43.2\%}$ top-1 test accuracy on CIFAR-10 dataset compared to state-of-art top-1 test accuracy $\textbf{39.8\%}$. }
\end{abstract}


%
\IEEEpeerreviewmaketitle

\section{Introduction}
As deep learning becomes dominant in many important areas, ensuring robustness becomes increasingly important. Deep neural networks are known to be vulnerable to adversarial attacks: small imperceptible perturbations in the inputs leading to incorrect decisions \cite{42503,biggio2013evasion}. Although many works have proposed heuristic defenses for training robust classifiers, they are often shown to be inadequate against adaptive attacks \cite{goodfellow2014explaining, carlini2017adversarial,uesato2018adversarial}. Therefore, a growing literature on certifiable robustness has emerged \cite{wong2018provable,cohen2019certified,zhang2024filtered,lecuyer2019certified,lyu2024adaptive,xiao2022densepure}; where the classifier's prediction \textit{must be provably robust} around any input  within a perturbation budget.

There are two broad approaches to design classifiers which are certifiably robust:  a) design classifiers which are inherently stable (i.e., smaller Lipschitz constants) \cite{gowal2018effectiveness, mirman2018differentiable, lee2020lipschitz}. There are a variety of methods to train classifiers while keeping the Lipschitz constants bounded. The second approach is b) randomized smoothing (RS) \cite{cohen2019certified, jeong2021smoothmix,lecuyer2019certified}; here, the idea is to smooth the decision of a base classifier by adding noise at the input. The approach of RS 
has been generalized in several directions: Salman et al. \cite{salman2020denoised} and Carlini et al. \cite{carlini2023certified} combine denoising mechanisms with smoothed classifiers, Salman et al. \cite{salman2019provably} combine adversarial training with smoothed classifiers, Zhai et al. \cite{Zhai2020MACER:} propose a regularization which maximizes the approximate certified radius and Horváth et al. \cite{horvath2021boosting} combine ensemble models with smoothed classifiers. 

Ensuring certified robustness by constraining the Lipschitz constant usually involves estimating the Lipschitz constant for an arbitrary neural network. The main challenge is that accurate estimation of Lipschitz constants becomes hard for larger networks, and upper bounds become loose leading to vacuous bounds on certified radius. Thus, a variety of approaches have emerged that focus on training while explicitly constraining the model's Lipschitz constant (outputs), for instance,  LMT \cite{tsuzuku2018lipschitz} and BCP \cite{lee2020lipschitz}. These methods control the Lischitz constant of the model by constraining the outputs of each layer (or the outputs of the model); this has the dual benefit of enhanced robustness of the model as well as better estimation of the overall Lipschitz constant of the model. To further minimize the Lipschitz constant during the certified robust training process and better estimate the \textit{local} Lipschitz constant of the model, recent work \cite{huang2021training} focus on the constrained training with respect to the \textit{local} Lipschitz constant by utilizing the clipped version of the activation functions. 

Another line of works \cite{xu2022lot,wang2020orthogonal} propose using models for which each individual layer is $1$-Lipschitz. By enforcing orthogonal or near-orthogonal weight matrices, these networks naturally limit their sensitivity to input perturbations, contributing to a form of robustness that does not solely depend on Lipschitz constant estimation. This approach can complement the limitations of both Lipschitz constrained training and RS, particularly in handling larger models where direct Lipschitz estimation becomes impractical. Orthogonal constraints help maintain the \textit{global} Lipschitz constant close to $1$. 

Lipschitz constrained training provides deterministic guarantees on certified radius and is often challenging to accurately estimate the Lipschitz constant of a large neural network.  RS on the other hand offers scalability to arbitrarily large networks and provide the closed-form certified robust radius. These guarantees, however, are probabilistic in nature and the smoothing procedure treats the entire classifier as a black box. 

\begin{table*}[t]
    \centering
    \begin{tabular}{l c c c | c c c |c c c}
        \hline
        & \multicolumn{9}{c}{Certified Test Accuracy at $\epsilon~(\%)$}\\
        & \multicolumn{3}{c}{MNIST} & \multicolumn{3}{c}{CIFAR-10}& \multicolumn{3}{c}{ImageNet}\\
         \cline{2-10}
         Method  & 1.50 & 1.75 & 2.00 & 0.5 & 0.75 & 1.0  &  1.5 & 2.0 & 3.0\\
         \hline
         \hline
         RS \cite{cohen2019certified} & 67.3 & 46.2 & 32.5 & 43.0 & 32.0 & 22.0 &  29.0 & 19.0 & 12.0\\ 
         MACER \cite{Zhai2020MACER:}& 73.0 & 50.0 & 36.0 & 59.0 & 46.0 & 38.0 &  31.0 & 25.0 & 14.0 \\
         Consistency \cite{jeong2020consistency} & 82.2 & 70.5 & 45.5 & 58.1 & 48.5 & 37.8 &  34.0 & 24.0& 17.0\\
         SmoothMix \cite{jeong2021smoothmix} & 81.8 & 70.7 & 44.9 & 57.9 & 47.7 & 37.2 &  38.0 & 26.0 & 20.0 \\
         DRT \cite{yang2021certified}  & 83.3 & 69.6 & 48.3 & 60.2 & 50.5 & 39.8 &   \textbf{39.8} &  30.4 & 23.2\\
         RS+OrthoNN \cite{wang2020orthogonal} & 70.1 & 49.7 & 33.2 &  45.9  & 28.9 & 19.2 & 28.4 & 16.2  & 10.0\\
         \hline
         \textit{\textbf{SPLITZ}}  & 80.2 & \textbf{71.3} & \textbf{62.3} & \textbf{63.2} & \textbf{53.4} & \textbf{43.2}& \textbf{38.6}  &  \textbf{31.2} & \textbf{23.9}\\
         \hline
    \end{tabular}
 \caption{\revision{Comparison of certified test accuracy (\%) on MNIST, CIFAR-10, and ImageNet under $\ell_2$ norm perturbation (\revisionfinal{see CIFAR-10 and ImageNet results in Section \ref{sec:evaluation}, and MNIST results in Appendix \ref{sec:addtional experimental_results}}). Each entry lists the certified accuracy using numbers taken from respective papers (RS results on MNIST follow from previous benchmark papers \cite{jeong2020consistency,jeong2021smoothmix}).}}
    \label{tab:mnist results}   
    \vspace{-10pt}
\end{table*}

\begin{figure*}[t]
\centering
\includegraphics[scale=0.35]{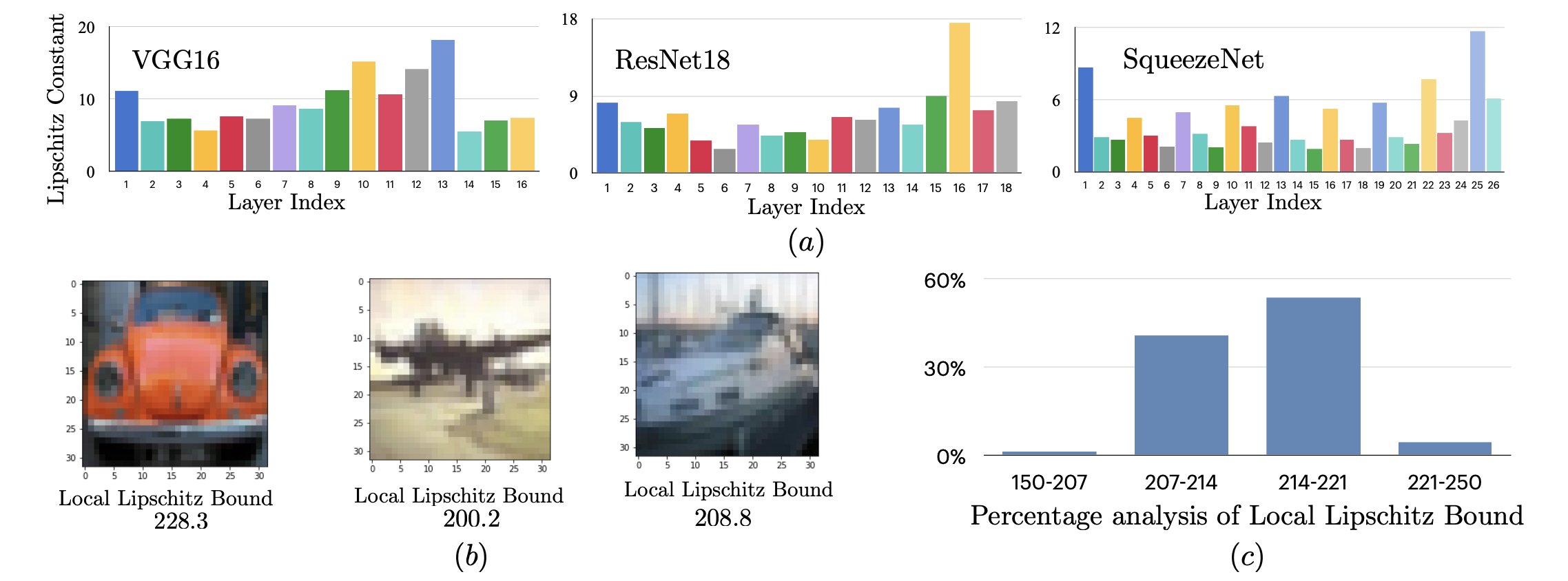}
\caption{(a) Lipschitz constants of each affine layer in pretrained models: VGG16 \cite{Simonyan15}, ResNet18 \cite{he2016deep}, SqueezeNet \cite{iandola2016squeezenet}. (b) Local Lipschitz (upper) bound for three random CIFAR-10 images on VGG16; (c) Percentage analysis of local Lipschitz (upper) bound in CIFAR-10 test data (additional results are presented in the Appendix \ref{Sec: addtional_Lipschitz_results}).}
\label{fig: Lip const}
\vspace{-10pt}
\end{figure*}

\vspace{10pt}
\noindent\textbf{Overview of \textit{SPLITZ} and Contributions}. In this paper, we propose \textit{SPLITZ}, which combines and leverages the synergies offered by both \textit{local} Lipschitz constrained training and randomized smoothing. The general idea is to split a classifier into two halves: the first half (usually a few layers) is constrained to keep a smaller Lipschitz (upper) bound, and the latter half of the network is \textit{smoothed} via randomization. We propose the use of \textit{local} Lipschitz constant(s) of the first half of the network. This is because it can capture the stability of the model with respect to each individual input. To enhance certified robustness, we incorporated the \textit{local} Lipschitz constant of the network's first half into the loss function as a regularization term. This approach aims to maintain a comparatively small \textit{local} Lipschitz constant (typically less than 1) for the network's first half, thereby improving the certified robustness. 
We provide the theoretical guarantee of \textit{SPLITZ} and derive a closed-form certified radius based on the \textit{local} Lipschitz constant as well as the randomized smoothing parameters as shown in Theorem \ref{the: general radius}; this result illustrates that enforcing a relatively small \textit{local} Lipschitz constant can help in improving the certified radius. To the best of our knowledge, this is the first systematic framework to combine the ideas of \textit{local} Lipschitz constants with randomized smoothing.

\revision{Interestingly, this approach yields state-of-the-art results for several datasets. For instance, Table \ref{tab:mnist results} compares the certified test accuracy of \textit{SPLITZ} and existing state-of-art techniques for various values of $\epsilon$ (perturbation budget or certified radius) on the MNIST, CIFAR-10 and ImageNet datasets. For $\epsilon$ as large as $2.0$, where the state-of-the-art accuracy is $48.3\%$, \textit{SPLITZ} achieves certified accuracy of around $62.3\%$. In Section \ref{sec:evaluation}, we present comprehensive set of results on MNIST, CIFAR-10 and ImageNet datasets. In addition, we also provide a detailed ablation study, and study the impact of various hyperparameters (such as the location of the split, randomized smoothing parameters, effects of local Lipschitz constant). }

\noindent\textbf{Intuition behind \textit{SPLITZ}}. The intuition behind \textit{SPLITZ} comes from the following key observations: 
a) \textit{Layer-wise Heterogeneity}: many modern deep networks exhibit heterogeneity in Lipschitz constants across layers. Fig \ref{fig: Lip const}(a) shows the per-layer Lipschitz constants for three networks (VGG16, ResNet18 and SqueezeNet). We observe that the values can vary widely across the layers, and quite often, latter half of the network often shows larger Lipschitz constants. b) \textit{Input (local) heterogeneity}: We show the local Lipschitz (upper) bounds for three randomly sampled images from CIFAR-10 when passed through the first four layers of VGG16; note that the values of local Lipschitz bound can vary across different inputs (images). The same behavior across the entire CIFAR-10 test dataset is shown in Fig. \ref{fig: Lip const}(c). These observations motivate \textit{SPLITZ} as follows: smoothing the input directly may not be the optimal approach as it does not account for this heterogeneity. Instead, by introducing noise at an intermediate stage of the classifier, the model can become more resilient to perturbations. This suggests the idea of splitting the classifier. Simultaneously, the first half of the network should also be ``stable", which motivates constraining the Lipschitz bound of first half of the network. 

The paper is organized as follows: Section \ref{sec: prelim} introduces the objectives of the paper, preliminaries, and definitions of randomized smoothing and Lipschitz constant(s) used in the paper. Additionally, we review related works concerning randomized smoothing and Lipschitz training. Section \ref{sec:splitz} discusses the theoretical guarantees of \textit{SPLITZ} and the corresponding training mechanisms. Experiments and evaluation results are presented and discussed in Section \ref{sec:evaluation}. Furthermore, additional experimental details on the Lipschitz constant, theoretical results on the local Lipschitz bound, and supplementary experimental results of the main findings are provided in the Appendices \ref{Sec: addtional_Lipschitz_results},  \ref{sec:Lipschitz_bound_theory}, and \ref{sec:addtional experimental_results}, respectively.
\section{Preliminaries on Certified Robustness} \label{sec: prelim}

We consider a robust training problem for multi-class supervised classification, where we are given a dataset of size $N$, $\{x_i, y_i\}_{i=1}^N$, 
where $x_i \in \mathbb{R}^d$ denotes the set of features of the $i$th training sample, and $y_i \in \mathcal{Y} := \{1,2,\dots,C\}$ represents the corresponding true label. We use $f$ to denote a classifier, which is a mapping $f$: $\mathbb{R}^d \rightarrow \mathcal{Y}$ from input data space to output labels.
From the scope of this paper, our goal is to learn a classifier which satisfies certified robustness, as defined next. 
\begin{definition}(Certified Robustness) \label{def: certified robustness}
A (randomized) classifier $f$ satisfies $(\epsilon, \alpha)$ certified robustness if for any input $x$, we have 
\begin{center}
    $\mathbb{P}(f(x) = f(x')) \geq 1- \alpha$, $\forall x'$, such that $ x' = x + \delta, ||\delta||_p \leq \epsilon$
\end{center}
where the probability above is computed w.r.t. randomness of the classifier $f$.
\end{definition}

Intuitively, certified robustness requires that for any test input $x$, the classifier's decision remains locally invariant, i.e., for all $\forall x'$ around $x$, such that $\parallel x'-x \parallel_{p} \leq \epsilon$, $f(x) = f(x')$ with a high probability. Thus, $\epsilon$ is referred to as the certified radius, and $(1-\alpha)$ measures the confidence. We mainly focus on $\ell_2$ norm ($p=2$) for the scope of this paper. 

The literature on certified robustness has largely evolved around two distinct techniques:  \textit{Randomized Smoothing} and \textit{Lipschitz constrained training for Certifiably Robustness}. We first briefly summarize and give an overview of these two frameworks, before presenting our proposed approach of \textit{Split Lipschitz Smoothing}. 

\vspace{10pt}
\noindent\textbf{Randomized Smoothing (RS)} \cite{cohen2019certified} is a general procedure, which takes an arbitrary classifier (base classifier) $f$, and converts it into a "smooth" version classifier (smooth classifier). Most importantly, the smooth classifier preserves nice certified robustness property and provides easily computed closed-form certified radius. Specifically, a general smooth classifier $g_{RS}(\cdot)$ derived from $f$ is given as: 
\begin{align}
    g_{RS}(x) =  \underset{c~\in~\mathcal{Y}}{\text{argmax}}~ \underset{\delta \sim \mathcal{N}(0,\sigma^2 I)}{\mathbb{P}}(f(x + \delta)=c)
\end{align}

Intuitively, for an input $x$, $g(x)$ will output the most probable class predicted by the base classifier $f$ in the neighbourhood of $x$ with a high confidence $1-\alpha$. In the paper \cite{cohen2019certified}, they prove that $g(x)$ is robust against $\ell_2$ perturbation ball of radius $\epsilon= \sigma \Phi^{-1}(\underline{p_A})$ around $x$, where $\sigma$ is the standard deviation of the Gaussian noise, and $\underline{p_A}$ is the lower bound of the probability that the most probable class predicted by the classifier $f$ is $c_A$.  
RS is arguably the only certified defense which can scale to large image classification datasets. Based on RS, a number of studies have been undertaken in this field: RS was originally proposed to deal with $\ell_2$ norm bounded perturbations; but was subsequently extended to other norms using different smoothing distributions, including $\ell_0$ norm with a discrete distribution \cite{lee2019tight}, $\ell_1$ norm with a Laplace distribution \cite{teng2020ell_1}, and the $\ell_{\infty}$ norm with a generalized Gaussian distribution \cite{zhang2020filling}.
Other generalizations include combining RS with adversarial training to further improve certified robustness and generalization performance \cite{salman2019provably} or denoising mechanisms (such as diffusion models) are often considered in conjunction with RS \cite{salman2020denoised,carlini2023certified}.

Achieving a \textit{large} certified radius can be equivalently viewed as learning a classifier with \textit{small} Lipschitz constant. The Lipschitz constant is a fundamental factor  in numerous studies focused on training a certifiably robust neural network, which can be defined as follows:

\begin{definition}(Global and Local Lipschitz Constant(s)) For a function $f$: $\mathbb{R}^{d}\rightarrow \mathcal{Y}$, the 
Global, Local, and $\gamma$-Local Lipschitz constants (at an input $x$)
 are respectively, defined as follows:
\begin{align}
&\text{(Global Lipschitz constant)~~} \nonumber\\
&~~~~~~L_f = \underset{x,y \in \text{dom}(f); x\neq y }{\sup} \frac{|| f(y)-f(x) ||_p}{|| y-x||_p}\\ 
&\text{(Local Lipschitz constant)~~}\nonumber\\ &~~~~~~L_f(x) = \underset{y \in \text{dom}(f); y\neq x }{\sup} \frac{|| f(y)-f(x) ||_p}{|| y-x||_p}\\
&\text{($\gamma$-Local Lipschitz constant)~~}\nonumber\\
&~~~~~~L^{(\gamma)}_f(x) = \underset{y \in B(x, \gamma); y\neq x }{\sup} \frac{|| f(y)-f(x) ||_p}{|| y-x||_p},
\end{align}

where $B(x, \gamma)$ denotes the  $\ell_p$-ball around $x$ of radius $\gamma$, i.e., $B(x, \gamma)= \{u: ||u-x||_p\leq \gamma\}$. 
\end{definition}
Informally, $L^{(\gamma)}_f(x)$ captures the stability of the function $f$ in the neighborhood of $x$, where the neighborhood is characterized by an $\ell_p$-ball of radius $\gamma$. 
\begin{figure*}[t]
\centering
\includegraphics[scale=0.5]{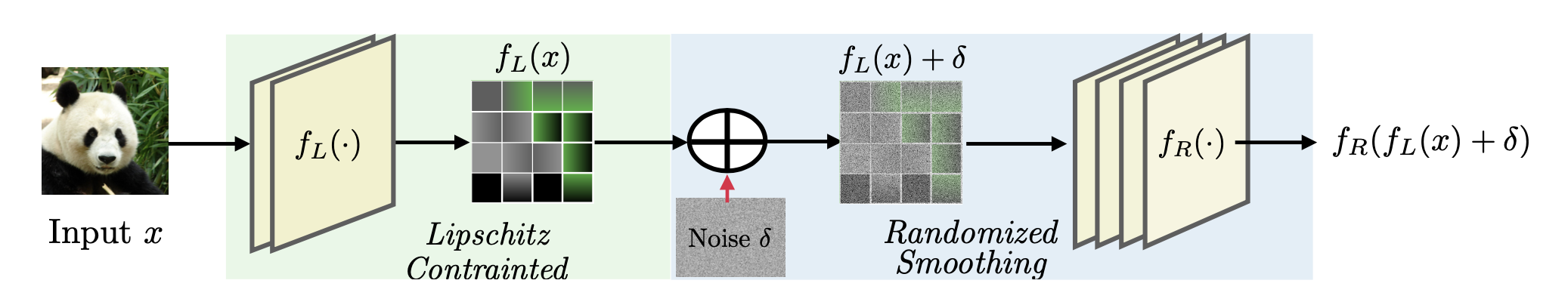}
\caption{Schematic of \textit{SPLITZ} training framework. We first feed the input $x$ to the left half of the classifier denoted as $f_{L}$, where the \textit{local} Lipschitz constant of $f_{L}$ is constrained. Subsequently, we smooth the right half of the classifier by introducing noise to the output of the left half, expressed as: $f_{L}(x) + \delta$. Finally, the output of the right half is $f_{R}(f_{L}(x) + \delta)$.}
\label{dp-compas-divergence}
\vspace{-5pt}
\end{figure*}

\vspace{10pt}
\noindent\textbf{Lipschitz constrained training for Certifiably Robustness}  A reliable upper bound for the local Lipschitz constant is essential for the robustness of a classifier. However, computing the exact value of local Lipschitz constants can be computationally challenging, prompting researchers to seek approximations, in terms of upper bounds. Thus, a line of works focus on deriving a tighter local Lipschitz bound e.g., \cite{zhang2019recurjac, fazlyab2019efficient, jordan2020exactly}. Another line of works utilize the local Lipschitz bound to obtain better robustness guarantees, e.g., \cite{hein2017formal, weng2018towards}. Furthermore, there are several works which aim to train a certified robust classifier as we briefly summarize next. One approach is to estimate/upper bound the global Lipschitz constant of the classifier
(during each training epoch) and use it to ensure robustness. For instance, \cite{tsuzuku2018lipschitz,lee2020lipschitz,leino2021globally} follow this general approach. The challenge is that the bounds on global Lipschitz constants can be quite large, and do not necessarily translate to improve certified robustness. An alternative approach is to use a local Lipschitz bound (for each individual input $x$), as in \cite{huang2021training} and then explicitly minimize the Lipschitz bound during the training process. For simplicity, we refer to the upper bound of the local Lipschitz constant as the ‘‘local Lipschitz constant''.

\section{\textit{SPLITZ}: Inference, Certification and Training}\label{sec:splitz}
\vspace{5pt}

In this Section, we first describe the details of the proposed \textit{SPLITZ} classifier along with the motivation as well as key distinctions from prior work. We then present new theoretical results on certified radius for \textit{SPLITZ}. Subsequently, we describe the training methodology for \textit{SPLITZ} as well as inference and computation of the certified radius. 
Suppose we are given a base classifier  $f:\mathbb{R}^d \rightarrow \mathcal{Y}$ which is a composition of $K$ functions. Consider an arbitrary ``split" of $f$ as $f(\cdot)= f_{\text{R}}(f_{\text{L}}(\cdot)) \triangleq f_{\text{R}}\circ f_{\text{L}}$. As an example, if the classifier has $K=2$ hidden layers, i.e., $f(x)= f_2(f_1(x))$, then there are $K+1=3$ possible compositions/splits:
$a) f_{\text{R}}= I, f_{\text{L}}=  f_{2}\circ f_1,
b) f_{\text{R}}= f_{2}, f_{\text{L}}= f_1,
c) f_{\text{R}}= f_{2}\circ f_1, f_{\text{L}}= I $, where $I$ represents the identity function.
\begin{definition}(\textit{SPLITZ} Classifier)\label{def:SPLITZ}
Let $f$ be a base classifier: $\mathbb{R}^d \rightarrow \mathcal{Y}$. Consider an arbitrary split of $f$ as  $f(\cdot) = f_{\text{R}}(f_{\text{L}}(\cdot))$.  We define the \textit{SPLITZ} classifier $g_{\textit{SPLITZ}}(\cdot)$ as follows:
\begin{align}
    g_{\textit{SPLITZ}}(x) = \underset{c~\in~\mathcal{Y}}{\arg\max}~\underset{\delta \sim \mathcal{N}(0,\sigma^2 I)}{\mathbb{P}}(f_{\text{R}}(f_{\text{L}}(x) + \delta)=c)
\end{align}
\end{definition}
The \textit{SPLITZ} smoothing classifier is illustrated in Fig \ref{dp-compas-divergence}. The basic idea of \textit{SPLITZ} is two fold:  smooth the \textit{right half} of the network using randomized smoothing and constrain the Lipschitz constant of the \textit{left half}. Specifically, to robustly classify an input $x$, we add noise to the output of the left half (equivalently, input to the right half) of the network, i.e., $f_{\text{L}}(x)$ and then follow the same strategy as randomized smoothing thereafter. While RS takes care of smoothing the right half, we would still like the left half to be as \textit{stable} as possible. Thus in addition to smoothing, we need to ensure that the Lipschitz constant of the left half $f_{\text{L}}$ of the network is also kept small. We next present our main theoretical result, which allows us to compute the certified radius for \textit{SPLITZ}.

\begin{theorem} \label{the: general radius}
    Let us denote $L_{f_\text{L}}^{(\gamma)}(x)$ as the $\gamma$-local Lipschitz constant of the function $f_\text{L}$ at $x$ in a ball of size $\gamma$, and $R_{f_\text{R}}(f_\text{L}(x))$ as the certified radius of the function $f_\text{R}$ at the input $f_\text{L}(x)$, with probability at least $(1-\alpha)$. Then, for any input $x$, with probability $1-\alpha$, $g_{\textit{SPLITZ}}(x)$ has a certified radius of at least,
    \begin{align}
        R_{g_{\textit{SPLITZ}}}(x) 
        = \underset{\gamma \geq 0 }{\max}~~\min \left\{\frac{R_{f_\text{R}}(f_\text{L}(x))}{L_{f_\text{L}}^{(\gamma)}(x)}, \gamma \right\}
    \end{align}
\end{theorem}
\begin{figure*}[t]
\centering
\includegraphics[scale=0.63]{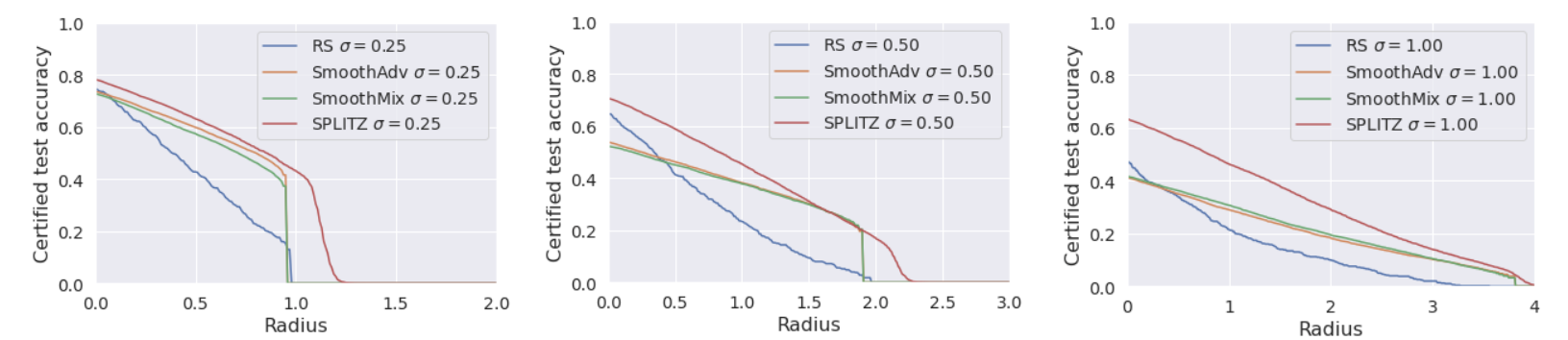}
\caption{Comparison of certified radius with $\ell_2$ norm perturbation w.r.t \textit{RS} \cite{cohen2019certified}, SmoothAdv \cite{salman2019provably}, Smoothmix \cite{jeong2021smoothmix} and \textit{SPLITZ} (ours), when varying levels of Gaussian noise $\sigma \in \{0.25, 0.5, 1.0\}$ on the CIFAR-10 dataset. We can observe that \textit{SPLITZ} consistently outperforms \textit{RS} \cite{cohen2019certified} under different noise levels. For instance, when Radius = 1.0, \textit{SPLITZ} achieves $43.2\%$ certified test accuracy while the best certified test accuracy of \textit{RS} under different noise levels is $22.0\%$. We refer the reader to Table \ref{tab:cifar results} for a comprehensive comparison of SPLITZ with several other approaches on CIFAR-10 dataset.}
\label{fig: overall compa}
\vspace{-15pt}
\end{figure*}
 Given an input $x$, to compute the certified radius for SPLITZ classifier, we need $L_{f_\text{L}}^{(\gamma)}(x)$, i.e., the $\gamma$-local Lipschitz constant (discussed in the next Section) and the certified radius of right half of the classifier, i.e., $R_{f_{\text{R}}}(f_{\text{L}}(x))$. For Gaussian noise perturbation in the second half,  $R_{f_{\text{R}}}(f_{\text{L}}(x))$ is exactly the randomized smoothing $\ell_2$ radius \cite{cohen2019certified}, given as
    $R_{f_{\text{R}}}(f_{\text{L}}(x)) = \frac{\sigma}{2}(\Phi^{-1}(\underline{p_A})-\Phi^{-1}(\overline{p_B}))$, where $\underline{p_A}$ is the lower bound of the probability that the most probable class $c_A$ is returned, $\underline{p_B}$ is upper bound of the probability that the “runner-up” class $c_B$ is returned.
    
\begin{remark}
\textbf{Optimization over $\gamma$} We note 
 from Theorem \ref{the: general radius} that finding the optimal choice of $\gamma$ is crucial. One way is to apply the efficient binary search during the certify process to find the optimal value of $\gamma$. Specifically, we set the initial value of $\gamma$ and compute the corresponding local Lipschitz constant $L_{f_L}^{(\gamma)}(x)$ at input $x$. By comparing the value between $\gamma$ and ${R_{f_R}(f_L(x))}/{L_{f_L}^{(\gamma)}(x)}$, we divide the search space into two halves at each iteration to narrow down the search space until 
 $\gamma^* = {R_{f_R}(f_L(x))}/{L_{f_L}^{(\gamma^*)}(x)}$. Another way is to do a one-step search. Specifically, we first approximate the local Lipschitz constant $\tilde L_{f_L}^{(\gamma)}(x)$ at $x$ by averaging local Lipschitz constants of inference data given the inference $\gamma$. We then set $\gamma' = {R_{f_R}(f_L(x))}/{\tilde L_{f_L}^{(\gamma)}(x)}$ and re-calculate the local Lipschitz constant $\tilde L_{f_L}^{(\gamma')}(x)$ according to $\gamma'$. Finally, we compute the approximate optimal $\gamma^* = {R_{f_R}(f_L(x))}/{\tilde L_{f_L}^{(\gamma')}(x)}$. \revision{ We study the behavior of the certified radius with respect to the value of $\gamma$ in Sec \ref{sec:evaluation} and we show that one step optimization is sufficient to achieve the desired certified radius with SPLITZ.}
Overall, we show the certification process in detail in Algorithm \ref{algo: LSTA}.   
\end{remark}


\begin{remark}
\textbf{Split Optimization} In Theorem \ref{the: general radius}, we presented our result for an arbitrary split of the classifier. In principle, we can also optimize over how we split the classifier.  If the base classifier is a composition of $K$ functions and the left part of the classifier $f_{L^{(s)}}$ contains $s$ layers and $f_R$ contains $(K-s)$ layers, then we can find the optimal split $s^{*}$ by
 varying $s$ from $0, 1,2,\ldots, K$.  We can observe that selecting $s=0$ corresponds to conventional randomized smoothing whereas $s=K$ corresponds to label smoothing. In our experiments (see Section \ref{sec:evaluation}), we find that it is sufficient to split after a few layers (e.g., split the classifier after the $s=1^{\text{st}}$ layer, $f_{\text{L}}= f_1$) and this alone suffices to outperform the state-of-art methods \cite{salman2019provably,cohen2019certified,jeong2020consistency,jeong2021smoothmix} on the CIFAR-10 dataset, where we show the comparison of certified radius in Fig \ref{fig: overall compa}. Similar behavior can also be observed on other image datasets such as the MNIST dataset. We further discuss the impact of different splitting strategies in Section \ref{sec:evaluation}. 
\end{remark}

\begin{remark}
\textbf{Compatibility of \textit{SPLITZ} with other defenses} In addition, the \textit{SPLITZ} mechanism is also compatible with other RS based certified robust techniques, such as adversarial smoothing \cite{salman2019provably}, mixsmoothing \cite{jeong2021smoothmix} or denoising diffusion models \cite{carlini2023certified}. For example, \cite{carlini2023certified} propose a denosing mechanism using a diffusion model, which achieves the state-of-the-art. Our \textit{SPLITZ} classifier contains two parts, left part is constrained by a small local Lipschitz constant while right part is smoothed by noise, which is same as a randomized smoothing based mechanism. Thus, our model can easily add a diffusion denoising model after the noise layer (after $f_L(x)+\delta$) and then feed the denoised samples into the $f_{R}$. Similarly, for adversarial smoothing or mixsmoothing, \textit{SPLITZ} is adaptable to feed either adversarial examples ($f_L(x')+\delta$) or mixup samples ($f_L(\tilde x)+\delta$) respectively to the right half of the classifier $f_R$. On the other hand, \textit{SPLITZ} is also compatible to other Lipschitz constrained based mechanisms (e.g., orthogonal
neural network based mechanism \cite{wang2020orthogonal,xu2022lot}). For instance, we can incorporate an orthogonal constraint regularization into the loss function to enforce orthogonality within the convolution layer throughout the training process. The results of integrating the orthogonal convolution neural network with \textit{SPLITZ} are meticulously detailed in Section \ref{sec:evaluation}. 
\end{remark}

We next present our proof of Theorem \ref{the: general radius} as follows.
\begin{proof}
    Let us consider an input $x$ to \textit{SPLITZ} classifier $g_{\textit{SPLITZ}}(\cdot)$ and define the following function
    \begin{align}
        \tilde g(u) \triangleq \underset{c \in \mathcal{Y}}{argmax}~ \mathbb{P}_{\delta}(f_R(u+\delta) = c).
    \end{align}
    We first note from Definition \ref{def:SPLITZ} that $g_{\textit{SPLITZ}}(x)$ can be written as
    $g_{\textit{SPLITZ}}(x)= \tilde g(f_{L}(x))$, where the function $\tilde g$ is the smoothed version of $f_{R}$. We are given that the smooth version $\tilde g$ has a certified radius of $R_{\tilde g}(u) \triangleq R_{f_{\text{R}}}(u)$ with probability at least $1-\alpha$. This is equivalent to the statement that
    for all $u'$ such that $||u-u'||_p\leq R_{f_{\text{R}}}(u)$, we have:
    \begin{align}
        \tilde g(u) = \tilde g(u'). 
    \end{align}
    We are also given $L_{f_\text{L}}^{(\gamma)}(x)$, the $\gamma$-local Lipschitz constant of the function $f_\text{L}$ at $x$ in a ball of size $\gamma$. This implies that for all $||x-x'||_p\leq \gamma$,
    \begin{align}
        || f_L(x) - f_L(x') ||_p \leq L_{f_L}^{(\gamma)}(x)|| x-x'||_p
    \end{align}
    Now, setting $u=f_{\text{L}}(x)$ and $u'= f_{\text{L}}(x')$, we obtain
    \begin{align}
       ||u-u'||_p= || f_L(x) - f_L(x') ||_p \leq L_{f_L}^{(\gamma)}(x)|| x-x'||_p.
    \end{align}
    Now observe that ensuring $g_{SPLITZ}(x)= g_{SPLITZ}(x')$ is equivalent to ensuring $\tilde g(f_{L}(x))=\tilde g(f_{L}(x'))$, which in turn is equivalent to $\tilde g(u)=\tilde g(u')$. Thus, if we ensure that 
    \begin{align}
        L_{f_L}^{(\gamma)}(x)|| x-x'||_p\leq R_{f_{\text{R}}}(f_{\text{L}}(x)) 
       \leftrightarrow ||x-x'||_p \leq \frac{R_{f_R}(f_L(x))}{L_{f_L}^{(\gamma)}(x)}
    \end{align}
then we have:
\begin{align}
   g_{SPLITZ}(x)= g_{SPLITZ}(x').
\end{align}
 However, we also note that:
 \begin{align}
     ||x-x'||_p\leq \gamma,
 \end{align}
 therefore the certified radius is given by:
 \begin{align}
     \min\left \{ \frac{R_{f_R}(f_L(x))}{L_{f_L}^{(\gamma)}(x)}, \gamma\right \}.
 \end{align}
  We finally note that the choice of $\gamma$ (size of the ball) was arbitrary, and we can pick the \textit{optimum} choice that yields the largest radius. This leads to the final expression for certified radius for \textit{SPLITZ}:
\begin{align}
        R_{g_{\textit{SPLITZ}}}(x) 
        = \underset{\gamma \geq 0 }{\max}~~\min \left\{\frac{R_{f_\text{R}}(f_\text{L}(x))}{L_{f_\text{L}}^{(\gamma)}(x)}, \gamma \right\}
    \end{align}
    and completes the proof of the Theorem.
\end{proof}
We show the inference and certification process of \textit{SPLITZ} in Algorithm \ref{algo: LSTA}.

\vspace{5pt}
 \noindent\textbf{Training Methodology for \textit{SPLITZ}}\label{sec:training_methodology}
In this Section, we present the details on training the  \textit{SPLITZ} classifier. The key to ensuring the certified robustness of the \textit{SPLITZ} classifier is to keep the local Lipschitz constant of the left half of the classifier $f_L$ \textit{small} while smoothing the right half of the classifier $f_R$. Let us denote $w_L$, $w_R$ as the training parameters of $f_L$ and $f_R$, respectively. 
We propose the following training loss function:
\begin{align}\label{eq: expect_loss}
    & \underset{w_L,w_R}{\text{min}}~\frac{1-\lambda}{N}\sum_{i=1}^N \mathbb{E}_{\delta}[\text{Loss}(f_R^{w_R}(f_L^{w_L}(x_i)+\delta),y_i)] \nonumber\\
    &+ \frac{\lambda}{N} \sum_{i=1}^N \text{max}(\theta, L_{f_L^{w_L}}^{(\gamma)}(x_i)),
\end{align}
where $\lambda \in [0,1]$ is a hyperparameter controlling the tradeoff between accuracy and robustness, $\theta$ is a learnable parameter to optimize the local Lipschitz constant, and $\text{Loss}(\cdot)$ is the loss function (e.g., cross entropy loss). \revisionfinal{Following the literature on
    randomized smoothing, we replace the expectation operators over the random variable $\delta$ by sampling $\delta_1,\delta_2,\dots,\delta_q$ noisy instances and computing their empirical mean. Specifically, we replace the expectation in Equation~\ref{eq: expect_loss} with an empirical average over these samples. This yields the following empirical loss function:}
\begin{align}\label{eq: final_loss}
    & \underset{w_L,w_R}{\text{min}}~\underbrace{ \frac{1-\lambda}{N}\sum_{i=1}^N \left(\frac{1}{Q}\sum_{q=1}^Q \text{Loss}(f_R^{w_R}(f_L^{w_L}(x_i)+\delta_q),y_i) \right)}_{\text{"Smoothing" loss for} f_R} \nonumber\\&+ \underbrace{\frac{\lambda}{N} \sum_{i=1}^N \text{max}(\theta, L_{f_L^{w_L}}^{(\gamma)}(x_i))}_{\text{Lipschitz regularization for} f_L}.
\end{align}
As illustrated in Eq \ref{eq: final_loss}, we first input the image $x_i$ to the left part of the classifier $f_L(x_i)$ and then add noise $\delta_q$, which forms the noisy samples $f_L(x_i)+\delta_q$. We then feed noisy samples to the right part of the classifier $f_R(f_L(x_i)+\delta_q)$ and obtain the corresponding prediction. Given the true label $y_i$, the loss (e.g., cross entropy) w.r.t $x_i$ can be computed. At the same time, the local Lipschitz constant of the left part of the classifier needs to be minimized. To this end, we propose a regularization term to the loss function, which controls the local Lipschitz constant of $f_L$.
In addition, we do not want the value of the local Lipschitz constant to become too small during the training process, which may lead to a poor accuracy. Therefore, we set a learnable Lipschitz constant threshold $\theta$ for local Lipschitz constant of $f_L$, and use  $\text{max}(\theta, L_{f_L}^{(\gamma)}(x_i))$ as the regularization term. 
 \begin{algorithm}[t]
    \caption{\textit{SPLITZ} Inference \& Certification}\label{algo: LSTA}
    \begin{algorithmic}
     \STATE \textbf{Input:} Test input $x$, classifier $f^*_L, f^*_R$, noise level $\sigma$, hyperparameter $\gamma$, confidence parameter $\alpha$, number of noise samples to predict top class  $n_{0}$, number of noise samples to estimate the lower bound of the probability of the top class $n_{1}$.
 
    \STATE \textbf{Output:} The certified radius $R_{g_{\textit{SPLITZ}}}(x)$ and corresponding prediction of the given input $x$.
    \end{algorithmic}
    \begin{algorithmic}[1]
    \STATE Run the SPLITZ classifier $n_0$ times: $\{f^*_{R}(f^*_L(x)+\delta_{i})\}_{i=1}^{n_0}$ using $n_0$ independent noise realizations. 
    \STATE Compute $\text{count}_{n_{0}}(i)= \text{\# of times class $i$ is the top class}$. 
    SPLITZ inference/prediction: $c_A = \arg \max_{i}  \text{count}_{n_{0}}(i)$

    \STATE Approximate the lower confidence bound $\underline{p_A}$ of the probability of the top class $c_A$ from $n_{1}$ independent runs of the SPLITZ classifier with confidence $1-\alpha$. 
    
    \IF{$\underline{p_A}>0.5$}
    \STATE Compute the certified radius of $f^*_{R}$: $R_{f^*_R}(f^*_L(x)) \leftarrow \sigma \Phi^{-1}(\underline{p_A})$
    \STATE Optimize $\gamma^*$ (discussed in Remark $1$) and calculate the corresponding local Lipschitz bound on $f^*_{L}$. (Eq. \ref{eq:local_lip_two}). 

    \STATE Compute the overall certified radius $R_{g_{\textit{SPLITZ}}}(x)$ at $x$ (as shown in Theorem \ref{the: general radius}): $R_{g_{\textit{SPLITZ}}}(x) 
        \leftarrow \min \left\{ R_{f^*_\text{R}}(f^*_\text{L}(x))/L_{f^*_\text{L}}^{(\gamma^*)}(x), \gamma^* \right\}$

    \STATE \textbf{Return} prediction $c_A$ and certified radius $R_{g_{\textit{SPLITZ}}}(x)$
    \ELSE
    \STATE \textbf{Return} abstain
    \ENDIF
\end{algorithmic}
\end{algorithm}

\vspace{5pt}
\noindent\textbf{Computing the Local Lipschitz bound}: We note that both \textit{SPLITZ} training as well as inference/certification require the computation of the local Lipschitz constant of the left half of the network, i.e., $f_L$.  The simplest approach would be to use a bound on the global Lipschitz constant of $f_L$. For example, if $f_L$ is composed of $s$ layers, with each layer being a combination of an affine operation followed by ReLU nonlinearity, then the following simple bound could be used:
\begin{align}
   L^{(\gamma)}_{f_L}(x)\leq|| W_{s}||_2\times || W_{s-1}||_2 \dots || W_{1}||_2, \nonumber
\end{align}
where $W_s$ is the weight matrix of layer $s$ and 
$|| W_{s}||_2$ denotes the corresponding spectral norm. 
However, this bound, while easy to compute turns out to be quite loose. More importantly, it does not depend on the specific input $x$ as well as the parameter $\gamma$. Fortunately, bounding the local Lipschitz constant of a classifier is an important and a well studied problem. There are plenty of mechanisms to estimate the local Lipschitz bound of $f_L$. In principle, our \textit{SPLITZ} classifier is compatible with these local Lipschitz bound estimation algorithms. From the scope of this paper, we use the local Lischitz constant constrained methodology proposed in  \cite{huang2021training} which leads to much tighter bounds on the local Lipschitz constant and maintain the specificity on the input $x$. 
Specifically, we apply the clipped version of activation layers (e.g. ReLU) to constrain each affine layer's output and obtain the corresponding upper bound (UB) and the lower bound (LB) for each affine layer, where the classifier is given an input $x$ around a $\gamma$ ball. We use an indicator function $I^v$ to represent index of the rows or columns in the weight matrices of each affine layer, which within the range from LB to UB. By multiplying each affine layer's weight matrix and each clipped activation layer' indicator matrix, the tighter local Lipschtz constant can be obtained.
 Assume $f_L$ network contains $s$-affine-layer neural network and each affine layer is followed by a clipped version of the activation layer,  (upper bound of) the local Lipschitz constant $L$ of $f_L$ around the input $x$ is:
\begin{align}
  &L_{f_L}^{(\gamma)}(x)\leq\nonumber\\
  &\parallel W_{s}I^v_{s-1}\parallel_2\times \parallel I^v_{s-1}W_{s-1}I^v_{s-2}\parallel_2 \dots \parallel I^v_{1}W_{1}\parallel_2, \label{eq:local_lip_two}
\end{align}
where $W_s$ is the weight matrix of layer $s$. \revisionfinal{The local Lipschitz constant \( L_{f_L}^{(\gamma)}(x) \) is computed as a product of spectral norms of weight matrices, modulated by activation-dependent indicator functions, which capture the network’s sensitivity to input perturbations. The bias parameters do not influence the Jacobian, as it is computed with respect to the input \( x \), and thus have no effect on Equation~\eqref{eq:local_lip_two}.
} We include more details with respect to local Lipschitz bound in Appendix \ref{sec:Lipschitz_bound_theory}.

\begin{algorithm}[t]\caption{\textit{SPLITZ} Training}\label{algo: new LSTA}  
\begin{algorithmic}
 \STATE \textbf{Input:} Training set $D_{\text{train}}=\{x_i, y_i\}_{i=1}^{N_{\text{train}}}$; noise level $\sigma$, training steps $T$, Lipschitz threshold $\theta$, training hyperparameter $\gamma$.
 
 \STATE \textbf{Output:} $f_L^*$, $f_R^*$.
 \end{algorithmic}
 \begin{algorithmic}[1]
    \FOR{$t = 0 , \dots, T-1$}

    \STATE Compute local Lipschitz constant of $f_L$: $L_{f_L}^{\gamma}(x) \leftarrow Cal\_ Lip~(f_L, x, \gamma)$.
    \STATE Sample noise $\delta$ and add it to outputs of $f_L$ to obtain noise samples: $f_L(x)+\delta$
    \STATE Feed the noise samples to $f_R$ network to get the corresponding predictions: $f_R(f_L(x)+\delta)$
    \STATE Set the local Lipschitz threshold $\theta$ and minimize the loss function in Eq. \ref{eq: final_loss}.
    \ENDFOR 
 \end{algorithmic}
 \begin{algorithmic}
\STATE\textbf{Function}~$\text{Cal\_}\text{Lip}$~($f_L, x, \gamma$)
 \end{algorithmic}

\begin{algorithmic}[1]
\STATE Compute the $UB_k$ and $LB_k$ for each layer $k$ in $f_L(x)$ given the perturbation $\gamma$ around input $x$
\STATE Compute the indicator matrix $I^v_k$  for each layer $k$ 
\STATE Compute the local Lipschitz constant $L_{f_L}^{(\gamma)}(x)$ (Eq. \ref{eq:local_lip_two})
\end{algorithmic}
\begin{algorithmic}
\STATE\textbf{Return} $L_{f_L}^{(\gamma)}(x)$
\end{algorithmic}
\end{algorithm}


\noindent\textbf{Summary of \textit{SPLITZ} Training Methodology} Overall, our training procedure is presented in Algorithm \ref{algo: new LSTA}.
 During the process of computing local Lipschitz constant of $f_L$, for each iteration, we feed the input to the classifier $f_L$ and calculate the LB and UB of outputs of each affine layer in $f_L$ given the input $x$ within a $\gamma$ ball. We then can calculate the indicator matrix $I^v$ and compute the spectral norm of the reduced weight matrix $\parallel I^v_{s}W_{s}I^v_{s}\parallel$ for each layer $s$ in $f_L$ using \textit{power iteration}. By multiplying the reduced weight matrix of each affine layer in $f_L$, we are able to arrive at the local Lipschitz constant of $f_L$. 
 Secondly, we smooth the right half of the neural network $f_R$ by sampling from Gaussian noise with zero mean and adding it to the output of $f_L$. Then we feed the noisy samples $f_L(x)+\delta$ to $f_R$ and obtain the corresponding loss. Next, we minimize the overall loss and backward the parameters to optimize the overall network $f$. Finally, we certify the base classifier $f$ to obtain the  classifier $g_{\textit{SPLITZ}}$ as shown in Algorithm \ref{algo: LSTA}.

\section{Experiments and Evaluation Results}\label{sec:evaluation}

\revision{\noindent In this section, we evaluate the \textit{SPLITZ} classifier on three datasets, MNIST\cite{lecun1998gradient}, CIFAR-10 \cite{krizhevsky2009learning} and ImageNet \cite{russakovsky2015imagenet}. We also present results on the Adult Income and Law school dataset in Appendix \ref{sec:addtional experimental_results}, demonstrating that SPLITZ achieves better trade-offs between robustness and accuracy on both tabular and image datasets.
We report the approximate certified test accuracy and certified radius of smoothed classifiers over full test datasets in MNIST and CIFAR-10 datasets and a subsample of 1,000 test data in ImageNet dataset. Same as previous works, we vary the noise level $\sigma \in \{0.25, 0.5, 1.0\}$ for the smoothed models on CIFAR-10 and ImageNet dataset, $\sigma \in \{0.25, 0.5, 0.75,  1.0\}$ for MNIST dataset. We certified the same noise level $\sigma$ during the inference time. To ensure a fair comparison with previous works, we provide the highest reported results from each paper for the corresponding above levels of noise magnitudes. To improve certified robustness, we utilize the tighter local Lipschitz bound introduced in \cite{huang2021training}. For three datasets, we use the same model as previous works \cite{cohen2019certified,carlini2023certified,jeong2021smoothmix,jeong2020consistency} (LeNet for MNIST, ResNet110 for CIFAR-10, ResNet50 for ImageNet). More experimental details are described in Appendix \ref{sec:addtional experimental_results}.

\vspace{5pt}
\noindent\textbf{Evaluation metric}
Our evaluation metric to measure the certified robustness of the smooth classifier is based on the standard metric proposed in \cite{cohen2019certified}: \textit{the approximate certified test accuracy}, which can be estimated by the fraction of the test dataset which CERTIFY classifies are correctly classified and at the same time corresponding radius are larger than radius $\epsilon$ without abstaining. Another alternative metric is to measure the \textit{average certified radius} (ACR) considered by \cite{Zhai2020MACER:}. We show that \textit{SPLITZ} consistently outperforms other mechanisms w.r.t ACR. For all experiments, we applied the $\ell_2$ norm input perturbation.

Our code is available at: \url{https://github.com/MeiyuZhong/SPLITZ-Codes.git}.
\vspace{5pt}
\noindent\textbf{\textit{SPLITZ} Methodology} 
For all datasets, we split the classifier after $1^{\text{st}}$ affine layer where the left half of the classifier contains one convolution layer followed by the clipped ReLU layer (See Appendix \ref{sec:Lipschitz_bound_theory}). For the ImageNet dataset, the only difference is that we remove the \textit{BatchNorm} layer after the $1^{st}$ affine layer and we replace the ReLU layer with the clipped ReLU layer in the first half of the network, which helps us obtain a tighter local Lipschitz bound of the first half of the classifier. The rest of the classifier is the same as original models (LeNet for MNIST, ResNet110 for CIFAR-10, ResNet50 for ImageNet).

\vspace{5pt}
\noindent\textbf{Dataset Configuration}  \revisionfinal{For the MNIST and CIFAR-10 dataset(s), we draw $n_{0} = 10^{2},~n_{1}=10^{5}$ noise samples to certify the smoothing model following \cite{cohen2019certified,carlini2023certified,jeong2021smoothmix}. For ImageNet dataset, we draw $n_{0} = 10^{2},~n_{1}=10^{4}$ noise samples to certify the smoothing model following \cite{cohen2019certified,carlini2023certified,jeong2021smoothmix}.
We set the Lipschitz threshold (see Sec \ref{sec:training_methodology}) as $\theta=0.5$. For local Lipschitz constrained training, we set tradeoff parameter $\lambda$ (see Sec \ref{sec:training_methodology}) evenly decrease from $0.8 - 0.4$, $0.7 - 0.5$ and $0.9 - 0.7$ respectively. We use one Nvidia P100 GPU to train the \textit{SPLITZ} model with batch size $512, 256, 128$ respectively. We apply Adam Optimizer for three datasets. For the MNIST dataset, we conduct training for 150 epochs and utilize an early-stop strategy to search for the optimal classifier over an additional 150 epochs. We set the initial learning rate as 0.001. The learning rate is decayed (multiplied by 0.1) by 0.1 at every 50 epochs (50th, 100th...). For the CIFAR-10 dataset, we train 200 epochs using the ResNet110 and use the early-stop strategy to search for the optimal classifier over an additional 200 epochs. Furthermore, we set the initial learning rate as 0.001 and final learning rate as $10^{-6}$. The learning rate starts to evenly decay at each epoch from the half of the training epochs. 
For the ImageNet dataset, we train 200 epochs for the ResNet50 and set the initial learning rate as 0.01. The learning rate starts to decay at each 40 epochs. We use Adam optimizer for all datasets. We report our more experimental details in Appendix \ref{sec:addtional experimental_results}.}}

\vspace{5pt}
\noindent\textbf{Baseline mechanisms}
\revisionfinal{We compare our method with various existing techniques proposed for robust training of smoothed classifiers, as listed below: (a) PixelDP \cite{lecuyer2019certified}: cerified robust training with differential privacy mechanism; (b) RS \cite{cohen2019certified}: standard randomized smoothing with the classifier trained with Gaussian augmentation; (c) SmoothAdv \cite{salman2019provably}: adversarial training combined with randomized smoothing; (d) MACER \cite{Zhai2020MACER:}: a regularization approach which maximizes the approximate certified radius; (e) Consistency \cite{jeong2020consistency}: a KL-divergence based regularization that minimizes the variance of smoothed classifiers $ f (x + \delta)$ across $\delta$; (f) SmoothMix \cite{jeong2021smoothmix}: training on convex combinations of samples and corresponding adversarial on smoothed classifier; (g) Boosting \cite{horvath2021boosting}: a soft-ensemble scheme on smooth training; (h) DRT \cite{yang2021certified}: a lightweight regularized training on robust ensemble ML models; (i) ACES \cite{horvath2022robust}: a selection-mechanism combined with a smoothed classifier; (j) DDS \cite{carlini2023certified}: a denoised diffusion mechanism combined with a smoothed classifier; (k) RS \cite{cohen2019certified} + OrthoNN \cite{wang2020orthogonal}: an orthogonal convolutional layer followed by a randomized smoothing model; (l) APNDC \cite{chen2024diffusion}: a novel diffusion classifier that acts as an ensemble of exact Posterior noised diffusion classifiers, while incurring no additional computational overhead.}
\begin{table}[t]
    \centering
    \begin{tabular}{l c c c c c}
        \hline
        CIFAR-10 & & \multicolumn{4}{c}{Certified accuracy at $\epsilon~(\%)$}\\
         \cline{3-6}
         Method &  Extra data & 0.25 & 0.5 & 0.75 & 1.0  \\
         \hline
         \hline
         PixelDP \cite{lecuyer2019certified} & \xmark
         & 22.0 & 2.0 & 0.0 & 0.0 \\
         
         RS \cite{cohen2019certified} & \xmark & 61.0 & 43.0 & 32.0 & 22.0\\
         SmoothAdv \cite{salman2019provably} & \xmark & 67.4 & 57.6 & 47.8 & 38.3 \\
         SmoothAdv 
  \cite{salman2019provably} & \cmark & 74.9 & 63.4 & 51.9 & 39.6 \\
         MACER \cite{Zhai2020MACER:}& \xmark & 71.0 & 59.0 & 46.0 & 38.0  \\
         Consistency \cite{jeong2020consistency}& \xmark & 68.8 & 58.1 & 48.5 & 37.8 \\
         SmoothMix \cite{jeong2021smoothmix}& \xmark  & 67.9 & 57.9 & 47.7 & 37.2\\
         Boosting \cite{horvath2021boosting}& \xmark & 70.6 & 60.4 & 52.4 & 38.8  \\
         DRT \cite{yang2021certified} & \xmark & 70.4 & 60.2 & 50.5 & 39.8  \\ 
         ACES \cite{horvath2022robust} & \xmark & 69.0 & 57.2 & 47.0 & 37.8\\
         DDS \cite{carlini2023certified} & \cmark & 76.7 & 63.0 & 45.3 & 32.1 \\
         DDS (finetuning) \cite{carlini2023certified} & \cmark& 79.3& 65.5& 48.7& 35.5\\
         APNDC \cite{chen2024diffusion}& \xmark & 82.2 &
        70.7 &
        \textbf{54.5 }&
        38.2\\
         RS + OrthoNN \cite{wang2020orthogonal} & \xmark & 63.3 & 45.9  & 28.9 & 19.2\\
        \hline
         \textit{\textbf{SPLITZ}}  & \xmark & 71.3 & 
         63.2 & \textbf{53.4} & \textbf{43.2}\\

         \hline
    \end{tabular}
    \caption{\revisionfinal{Comparison of the approximate certified test accuracy (\%) on CIFAR-10 under $\ell_2$ norm perturbation. Extra data indicates whether their models incorporate other datasets in their models. Each entry lists the certified accuracy using numbers taken from respective papers.}}
    \label{tab:cifar results}
    \vspace{-10pt}
\end{table}
\subsection{Main Results}\label{sec:Results}

\vspace{5pt}
\noindent\textbf{Results on CIFAR10} As shown in Table \ref{tab:cifar results} and Fig \ref{fig: overall compa}, our method outperforms the state-of-art approaches for every value of $\epsilon$ on CIFAR-10 dataset. 
Interestingly, we find that the \textit{SPLITZ} training has a significant improvement when the value of $\epsilon$ is large. For instance, when $\epsilon = 1.0$, the model achieves $\textbf{43.2\%}$ top-1 test accuracy on CIFAR-10 dataset compared to state-of-art top-1 test accuracy 39.8\%. One hypothesis is that minimizing the Lipscitz bound of $f_L$ ($L_{f_L} \leq 1$) is able to boost the certified radius of the model. Intuitively, more samples are correctly classified while corresponding radius are larger than given $\epsilon$. In addition, we can observe the similar trend as MNIST dataset. \textit{SPLITZ} maintains higher certified test accuracy when we increase $\epsilon$ from 0.25 to 1.00 compared to other state-of-art mechanisms.

\begin{table}[t]
    \centering
    \begin{tabular}{l c c c c c  }
        \hline
         ImageNet & & \multicolumn{4}{c}{Certified accuracy at $\epsilon~(\%)$}\\
         \cline{3-6}
         Method &  Extra data & 1 & 1.5 & 2.0  & 3.0 \\
         \hline
         \hline
         PixelDP \cite{lecuyer2019certified} & \xmark
         & 0.0 & 0.0 & 0.0 &  0.0 \\
         
         RS \cite{cohen2019certified} & \xmark  & 37.0 & 29.0 & 19.0 &  12.0 \\
         SmoothAdv \cite{salman2019provably} & \xmark  & 43.0 &  37.0 & 27.0 & 20.0 \\
         MACER \cite{Zhai2020MACER:}& \xmark  & 43.0 & 31.0 & 25.0 &  14.0  \\
         Consistency \cite{jeong2020consistency}& \xmark  & 44.0 & 34.0 & 24.0 & 17.0 \\
         SmoothMix \cite{jeong2021smoothmix}& \xmark & 43.0 & 38.0 & 26.0 &  20.0\\
         Boosting \cite{horvath2021boosting}& \xmark & 44.6 & 38.4 & 28.6 &  21.2  \\
         DRT \cite{yang2021certified} & \xmark & 44.4 &  \textbf{39.8} &  30.4 &  23.2  \\ 
         ACES \cite{horvath2022robust} & \xmark & 42.2 & 35.6 & 25.6 & 19.8\\
          DDS \cite{carlini2023certified} & \cmark  & 54.3 & 38.1 & 29.5 & 13.1 \\
         RS + OrthoNN \cite{wang2020orthogonal} & \xmark & 38.0  & 28.4 & 16.2 & 10.0 \\
         \hline
         \textit{\textbf{SPLITZ}}  & \xmark &  43.2 & \textbf{38.6}  &  \textbf{31.2} & \textbf{23.9}\\
         \hline
    \end{tabular}
    \caption{\revision{Comparison of the approximate certified test accuracy (\%) on ImageNet under $\ell_2$ norm perturbation. The columns and rows have the same meaning as in Table \ref{tab:cifar results}.}}
    \label{tab:imagent}
    \vspace{-10pt}
\end{table}
\begin{table}[t]
    \centering

    \scalebox{1.0}{
    \begin{tabular}{l l c c c}
       \hline
         $\sigma$ & Methods & MNIST & CIFAR-10 & ImageNet  \\
        \hline
        \hline
         \multirow{6}{4em}{0.50} & $\text{RS}^{*}$ &1.553 &0.525&  0.733 \\
         & $\text{SmoothAdv}^{*}$ &1.687&  0.684& 0.825 \\
         & MACER & 1.583 & 0.726 & 0.831 \\
         & Consistency&1.697 &0.726& 0.822\\
         & SmoothMix &  1.694 & 0.737 &  0.846 \\

         & \textit{\textbf{SPLITZ}}  &\textbf{2.059} & \textbf{0.924} &\textbf{0.968} \\
         \hline
         \multirow{6}{4em}{1.00} & $\text{RS}^{*}$ & 1.620 & 0.542 &  0.875 \\
         & SmoothAdv & 1.779 & 0.660 & 1.040\\
         & MACER & 1.520 & 0.792 & 1.008 \\
         & Consistency& 1.819 & 0.816 & 0.982 \\
         & SmoothMix & 1.823 & 0.773 &  1.047 \\
         & \textit{\textbf{SPLITZ}}  &\textbf{2.104} & \textbf{0.979} & \textbf{1.282} \\
         \hline
    \end{tabular}}
        \caption{\revision{Comparison of average certified radius (ACR) across three different datasets (MNIST, CIFAR-10 and ImageNet). We can observe that for two datasets, \textit{SPLITZ} consistently achieves better results compared to other state-of-art mechanisms. * is reported by \cite{jeong2020consistency, jeong2021smoothmix}}}
    \label{tab:ACR results}
    \vspace{-10pt}
\end{table}

\vspace{5pt}
\revision{\noindent\textbf{Results on ImageNet}
We show the comparison of different certified robustness techniques on ImageNet dataset in Table \ref{tab:imagent}. We observe similar trends to MNIST and CIFAR10 datasets, where \textit{SPLITZ} is effective on certified robustness with a wide range of image datasets. For instance, when $\epsilon =2.0$, SPLITZ achieve $31.2\%$ while the state-of-the-art have $30.4\%$. Moreover, \textit{SPLITZ} consistently achieves better ACR (average certified radius) than other mechanisms in the ImageNet dataset, as shown in Table \ref{tab:ACR results}.

\vspace{5pt}
\noindent\textbf{Results on Average Certified Radius (ACR)}
We investigate the performance of \textit{SPLITZ} using average certified radius (ACR), where we measure the correct samples' average certified radius over the test datasets (MNIST, CIFAR-10, ImageNet). As shown in Table \ref{tab:ACR results}, we provide the comprehensive comparison results of average certified radius (ACR) compared to other certified robust techniques. Our \textit{SPLITZ} consistently outperforms others. For instance, when $\sigma = 0.50$, the ACR of \textit{SPLITZ} is 2.059 respectively, where the state-of-the-art is 0.933 in MNIST dataset. In addition, we can observe the same trends in CIFAR-10 and ImageNet dataset, where \textit{SPLITZ} consistently outperforms the state-of-the-art when $\sigma = 0.50, 1.00$.

\begin{table}[t]
    \centering
    \scalebox{1.1}{
    \begin{tabular}{ l c c c }
    \hline
      \multicolumn{4}{c}{FGSM}  \\
       \hline
          Methods & $\epsilon = 0.5$ & $\epsilon = 1.0$ & $\epsilon = 2.0$ \\
        \hline
        \hline
          RS &46.86 & 38.79  & 25.54 \\
          SPLITZ & \textbf{59.96} & \textbf{54.69} & \textbf{46.77} \\
         \hline
            \multicolumn{4}{c}{PGD }     \\
             Methods & $\epsilon = 0.5$ & $\epsilon = 1.0$ & $\epsilon = 2.0$ \\
        \hline
        \hline
         RS & 33.35  & 23.63  & 10.18 \\
         SPLITZ & \textbf{50.53} & \textbf{39.05} & \textbf{27.03} \\
         \hline
          \multicolumn{4}{c}{Auto-attack}\\
         Methods & $\epsilon = 0.5$ & $\epsilon = 1.0$ & $\epsilon = 2.0$ \\
        \hline
        \hline
         RS & 32.39  & 21.90 & 7.72 \\
         SPLITZ & \textbf{49.35}
         & \textbf{33.21} & \textbf{23.15} \\
         \hline
    \end{tabular}}
        \caption{\revision{Comparison of empirical certified test accuracy under attacks with respect to CIFAR-10 dataset.}}
    \label{tab:Emprical_results}
\end{table}}
\vspace{5pt}
\noindent\textbf{Empirical Robust Test Accuracy} \revisionfinal{We study the empirical robust test accuracy in Table \ref{tab:Emprical_results} under $\ell_2$ FGSM~\cite{goodfellow2014explaining}, PGD~\cite{madry2018towards}, and AutoAttack~\cite{croce2020reliable} using RS and \textit{SPLITZ} model with variance $\sigma =1$. Specifically, FGSM perturbs the input in the direction of the gradient sign to generate one-step adversarial examples; PGD extends this by iteratively applying FGSM with projection onto an $L_2$-ball to enforce bounded perturbations; AutoAttack is a parameter-free ensemble of strong white-box and black-box attacks that provides reliable robustness evaluation. } We can observe that the \textit{SPLITZ} consistently outperforms the RS method. For example, when $\epsilon = 2.0$ under auto-attack, \textit{SPLITZ} achieve the empirical test accuracy $23.15\%$ while RS is $7.72\%$.
\subsection{Training and certifying time} Our \textit{SPLITZ} model needs to vary the value of $\gamma$ (the size of the ball around input $x$) during the training epoch. Thus, we may need relatively more time to obtain the optimal model. To solve this, we apply the early stop mechanism to obtain a better optimized model during the training process. At the same time, we decay our learning rates during the training process. \revisionfinal{We report the training time of \textit{SPLITZ} and other baselines using one NVIDIA P100 GPU for CIFAR-10 and four NVIDIA P100 GPUs for ImageNet in the Table \ref{tab:training_time}. We observe that \textit{SPLITZ} requires slightly more training time than RS, but significantly less time than SmoothAdv and MACER.}


\subsection{Ablation Study}
We also conduct an ablation study to explore the effects of hyperparameters in our proposed method on CIFAR-10 and MNIST datasets. We will explain the impact of the splitting location, the effect of global (local) Lipschitz bound, comparison of the orthogonal neural network (OrthoNN) and \textit{SPLITZ}, effect of input perturbation $\gamma$ and effect of learnable Lipschitz threshold parameter $\theta$.


\begin{table}[t]
    \centering
    \scalebox{0.9}{
    \begin{tabular}{c c c }
    \hline
        Datasets &  Method & Training per epoch (s)  \\
        \hline
        \hline
        \multirow{4}{5em}{CIFAR-10} 
         & RS & 31.4 \\
         & SmoothAdv & 1990.1 \\
         & MACER & 504.0 \\
         & \textit{SPLITZ} & 59.5 \\
         \hline
        \multirow{4}{5em}{ImageNet} 
                 & RS & 2154.5 \\
         & SmoothAdv & 7723.8 \\
         & MACER & 3537.1 \\
         & \textit{SPLITZ} & 3160.5 \\
        \hline
        \hline
    \end{tabular}}
    \caption{\revisionfinal{Comparison of \textit{SPLITZ} training time across two datasets (CIFAR-10 using ResNet 110 and ImageNet using ResNet50) with RS \cite{cohen2019certified}, SmoothAdv \cite{salman2019provably} and MACER \cite{Zhai2020MACER:}. }}
    \label{tab:training_time}
    \vspace{-10pt}
    \end{table}

\begin{table}[t]
    \centering

    \scalebox{0.9}{
    \begin{tabular}{l c c c c c c c}
       \hline
        & \multicolumn{6}{c}{Certified Test Accuracy at $\epsilon~(\%)$}\\
        \cline{2-8}
         Location of Splitting & 0.50 & 0.75 & 1.00 & 1.25 & 1.50 & 1.75 & 2.00\\
        \hline
        \hline
        $1^{\text{st}}$ affine layer & \textbf{94.1} & \textbf{92.0} & \textbf{88.8} & \textbf{84.7} & \textbf{79.0} & \textbf{71.3} & \textbf{62.3} \\ 
         $2^{\text{nd}}$ affine layer & 
         89.7 & 86.2 & 81.0 & 75.0 & 68.0 & 59.9 & 51.0  \\ 

         $3^{\text{rd}}$ affine layer & 
         84.4 & 80.4 & 75.6 & 69.9 & 63.6 &
 56.9 & 49.4  \\ 

        \hline
        \hline
    \end{tabular}}
    \caption{Comparison of certified test accuracy of $\textit{SPLITZ}$ with Gaussian noise $\sigma = 0.75$ for varying the splitting layer on MNIST dataset with LeNet.}
    \label{tab:effect of the splitway}
\end{table}

\begin{figure*} 
    \centering
    \includegraphics[scale=0.25]{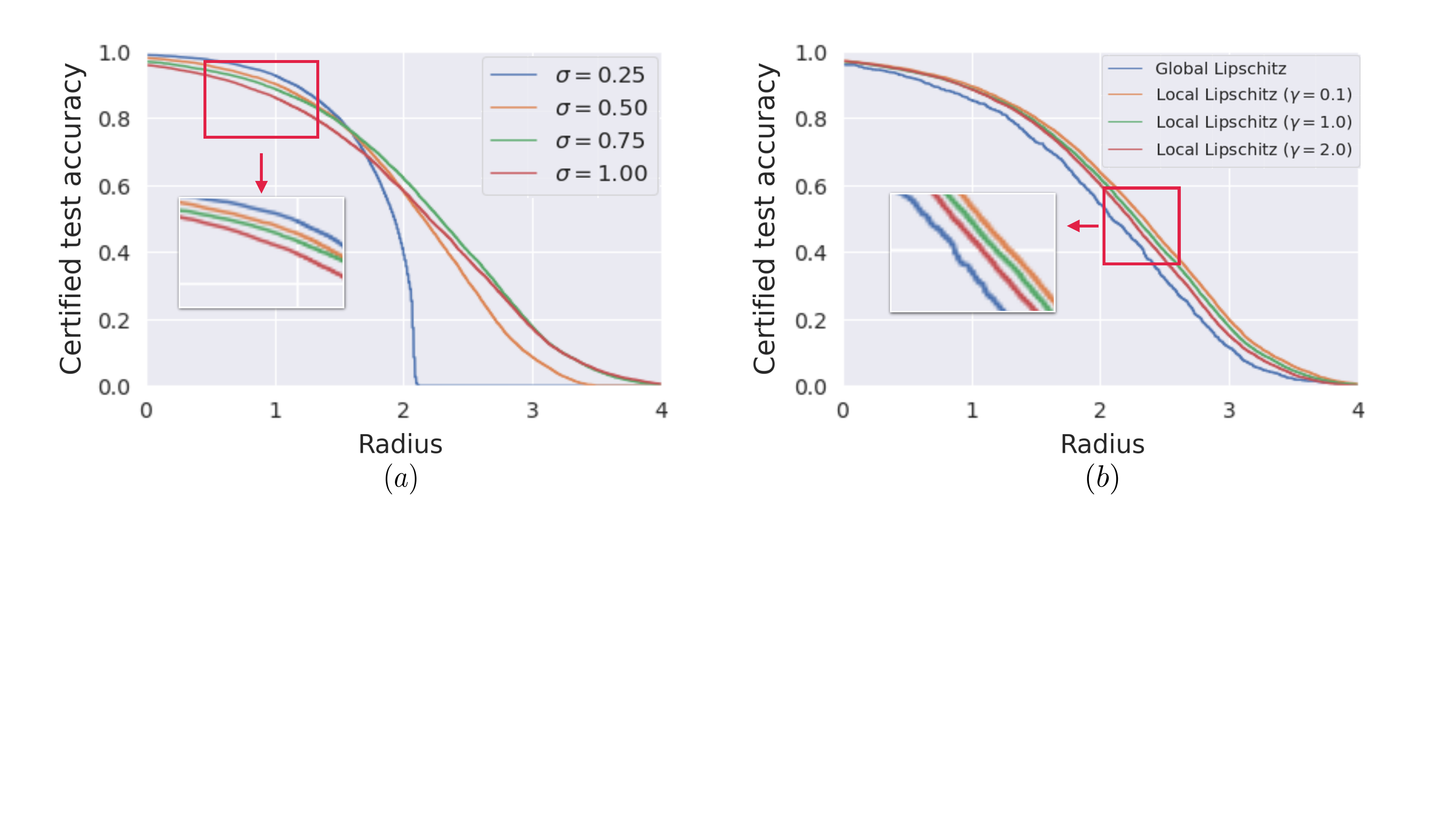}
    \caption{\revisionfinal{(a) Comparison of certified test accuracy of $\textit{SPLITZ}$ when varying $\sigma$ on the MNIST dataset, (b) Comparison of certified test accuracy of $\textit{SPLITZ}$ with Global vs Local Lipschitz bound ($\gamma$) on the MNIST dataset.} }
    \label{fig:ablation_global_thres}
    \vspace{-10pt}
\end{figure*}

\vspace{5pt}
\noindent\textbf{Impact of splitting location} As mentioned in Section \ref{sec:splitz}, our \textit{SPLITZ} classifier can be optimized over different split ways, where we conduct the experiments and show our results in Table \ref{tab:effect of the splitway}. For example, when $\epsilon = 2$, splitting after the $1^{st}$, or $2^{nd}$, or $3^{rd}$ layer result in certified accuracy of $62.3$\%, $51.0$\% and $49.4$\% respectively.  These results indicate that splitting the neural network early achieves better performance. Intuitively, splitting the neural network early helps the model minimize the local Lipschitz bound, which improves the certified robustness leading to a higher certified test accuracy given the same $\epsilon$. As the splitting becomes "deeper", estimating the local Lipshitz constant also becomes harder, which implies that a looser bound leads to smaller certified radius. 

\vspace{5pt}
\noindent\textbf{Effect of global (local) Lipschitz constant of the first half of the classifier} As shown in Fig \ref{fig:ablation_global_thres} (a), we investigate the effect of (upper bound of) the Lipschitz constant of left half of the classifier on certified test accuracy. Interestingly, we can observe that tighter Lipschitz bound gives better certified accuracy given the same radius. Furthermore, using a bound on the local Lipschitz constant to compute the certified accuracy is always better than using the global Lipschitz constant. This is also clearly evident from the result of Theorem \ref{the: general radius}.

\vspace{5pt}
\noindent\textbf{Comparison of the orthogonal neural network (OrthoNN) and the \textit{SPLITZ}} In the baseline RS+OrthoNN model, during the training process, we enforce the first convolution layer of the classifier's left half to act as an orthogonal convolution neural network. This is achieved by incorporating an orthogonal constraint regularization into the loss function. For the \textit{SPLITZ} model, we impose a constraint on the local Lipschitz constant of the classifier's left half. \revisionfinal{The main advantage of SPLITZ is that it splits the neural network into two halves and constrains the Lipschitz constant of the left part to be less than 1 (instead of exactly 1 while using OrthoNN), which can boost the certified radius and enables a better trade-off between robustness and accuracy.} The comparison between RS+orthoNN and SPLITZ is illustrated in Table \ref{tab:mnist results_2} with respect to the MNIST dataset, Table \ref{tab:cifar results} with respect to the CIFAR-10 dataset and Table \ref{tab:imagent} with respect to the ImageNet dataset. Our observations indicate that enforcing an orthogonal convolution layer on the classifier's left half has a comparatively lesser impact than training with a local Lipschitz constraint.
\begin{table}[t]
    \centering
    \scalebox{0.9}{
    \begin{tabular}{c c c c c c c c c c   }
       \hline
        & \multicolumn{9}{c}{Certified Test Accuracy at $\epsilon~(\%)$}\\
        \cline{2-10}
         $Lip_\theta$ & 0.50 & 0.75 & 1.00 & 1.25 & 1.50 & 1.75 & 2.00 & 2.25 & 2.50  \\
        \hline
        \hline
         0.3 & 94.1 & 91.6 & 88.4 & 84.2 &78.2 & 70.3 & 60.9 & 50.6 & 39.4 \\
         0.5 & 94.1 & 92.0 & 88.8 & 84.7 & 79.0 & 71.3 & 62.3 & 51.7 & 40.5  \\
        0.7 & 94.7 & 92.6 & 89.4 & 85.1 & 79.5 & 71.5 & 61.7 & 50.7 & 38.8  \\    
         \hline
    \end{tabular}}
    \caption{Comparison of certified test accuracy of $\textit{SPLITZ}$ with Gaussian noise $\sigma = 0.75$ for varying $Lip_\theta$ (the threshold of the local Lipschitz bound around input $x$) on MNIST dataset. }
    \label{tab:effect of lip}
    \vspace{-10pt}
\end{table}

\begin{table}[t]
    \centering
    \scalebox{0.95}{
    \begin{tabular}{c c c c c c c c c c  }
       \hline
        & \multicolumn{9}{c}{Certified Test Accuracy at $\epsilon~(\%)$}\\
        \cline{2-10}
         $\gamma$  & 0.50 & 0.75 & 1.00 & 1.25 &  1.50 & 1.75 & 2.00 & 2.25 & 2.50  \\
        \hline
        \hline
         0.1 & 94.7 & 92.5 & 89.6 & 85.4 & 80.2 & 73.2 & 64.0 & 54.1 & 42.7\\
         1 & 94.1 & 92.0 & 88.8 & 84.7 & 79.0 & 71.3 & 62.3 & 51.7 & 40.5  \\
        2 & 94.3 & 92.1 & 88.6 & 84.2 & 78.3 & 70.4 & 60.5 &  49.6 & 38.0  \\    
         \hline
    \end{tabular}}
    \caption{Comparison of certified test accuracy of $\textit{SPLITZ}$ with Gaussian noise $\sigma = 0.75$ for varying $\gamma$ (the size of the ball around input $x$) on MNIST dataset.}
    \label{tab:effect of lip_gamma}
    \vspace{-10pt}
\end{table}
\begin{figure*}
    \centering
    \includegraphics[width=0.9\linewidth]{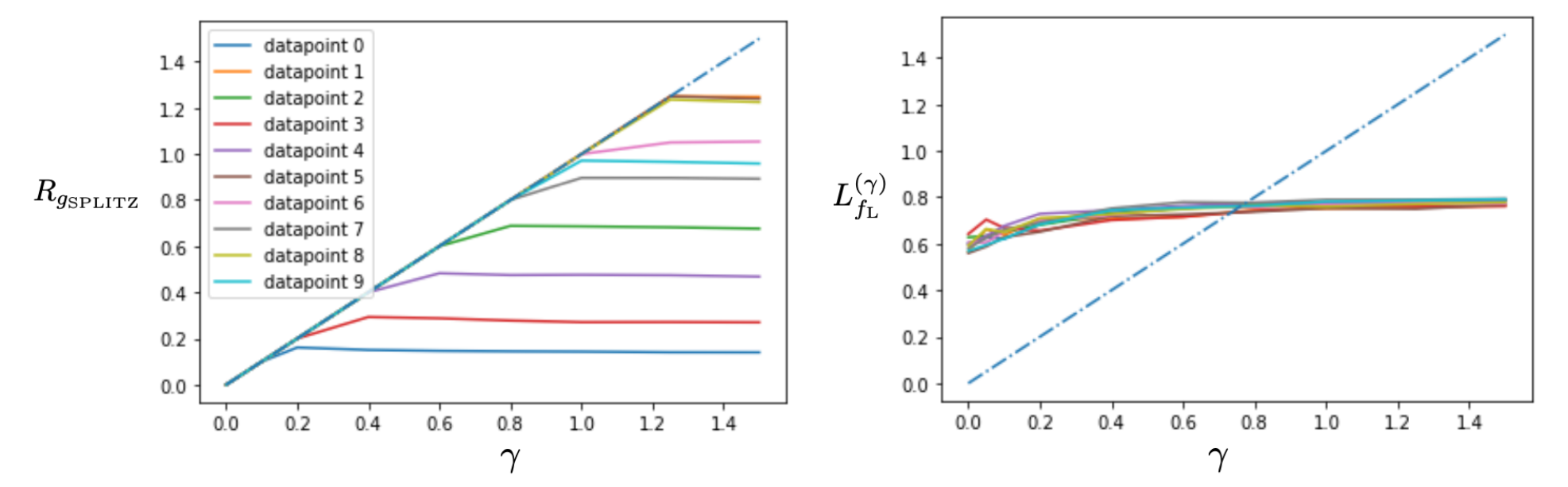}
    \caption{\revision{Left: Certified Radius of SPLITZ method vs the size of the ball around the input $\gamma$. Right: Local Lipschitz Constant vs the size of the ball around the input $\gamma$.}}
    \label{fig:optimization_over_gamma}
\end{figure*}

\vspace{5pt}
\noindent\textbf{Effect of $\theta$ (Lipschitz threshold)} 
As shown in Fig \ref{fig:ablation_global_thres}(b) and Table \ref{tab:effect of lip}, we analyze the effect of the training threshold $\theta$ (See Eq \ref{eq: final_loss}). For smaller values of $\epsilon$, \textit{SPLITZ} with higher Lipschitz constant achieves better performances on accuracy. Conversely, \textit{SPLITZ} with a smaller Lipschitz constant can boost certified radius, which obtains a relative higher certified test accuracy when $\epsilon$ is larger. These observations suggest a trade-off between the Lipschitz constant and certified accuracy. Thus, identifying the optimal Lipschitz threshold is essential for achieving  a balances between robustness and utility. This ablation study further validates that the essence of our \textit{SPLITZ} classifier lies in maintaining a relatively optimized Lipschitz constant for the left half of the classifier.

\vspace{5pt}
\noindent\textbf{Effect of $\gamma$ (size of radius around input $x$)} According to above results, constraining the local Lipschitz constant achieves better performance. To further explore the benefit of local Lipschitz constrained training, it is necessary to explore the indicator matrix $I^v$ in Eq \ref{eq:local_lip_two}, which depends on the size of the ball around the input (i.e., the hyperparameter $\gamma$). In Table \ref{tab:effect of lip_gamma}, we show how varying training $\gamma$ impacts the certified test accuracy for different values of $\epsilon$. 
We observe that smaller values of training $\gamma$ lead to higher certified accuracy for all values of $\epsilon$.

\vspace{5pt}
\revision{\noindent\textbf{Effect of $\gamma$ and local Lipshitz constant}
In this section, we study the optimization of the choice of $\gamma$. We show the tradeoffs between the certified radius of SPLITZ and the value of $\gamma$ in Fig \ref{fig:optimization_over_gamma} (Left). We randomly select 10 datapoints from CIFAR-10 dataset and certify them using the SPLITZ model with variance of 0.25. We also show the tradeoffs between the local Lipschitz constant and the value of $\gamma$. We observe that both the local Lipschitz constant and the the certified radius stabilizes after a certain increase in $\gamma$. Therefore, in practice, a one-step optimization is sufficient to obtain the certified radius of SPLITZ. However, if the optimal certified radius is desired, a binary search can be performed, although it may be more time-consuming than one-step optimization.}
\subsection{Results on Tabular Dataset}

\begin{figure}
    \centering
    \includegraphics[width=0.95\linewidth]{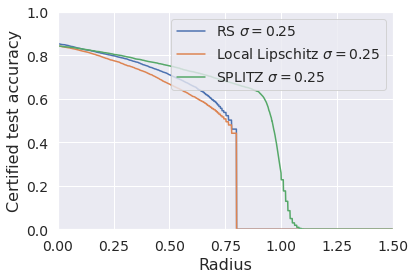}
    \caption{\revision{Tradeoffs between the certified robustness and accuracy with respect to the RS \cite{cohen2019certified}, local Lipschitz constant constrained \cite{huang2021training} and SPLITZ method on the Adult Income Dataset.}}
    \label{fig:adult_cd}
\end{figure}
\begin{figure}
    \centering
    \includegraphics[width=0.95\linewidth]{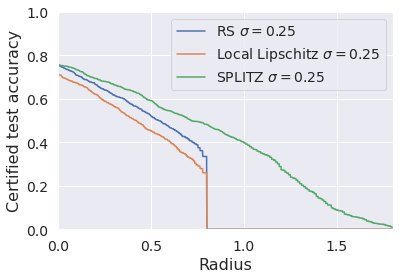}
    \caption{\revision{Tradeoffs between the certified robustness and accuracy with respect to the RS \cite{cohen2019certified}, local Lipschitz constant constrained \cite{huang2021training} and SPLITZ method on the Law School Dataset.}}
    \label{fig:law_cd}
\end{figure}

 \revision{\noindent\textbf{Results on Adult Income Dataset} Adult income dataset includes income related data with 14 features (i.e., age, work class, occupation, education etc.) of $N=45,222$ users ($N_{train} = 32,561$, $N_{test} = 12,661$) to predict whether the income of a person exceeds a threshold (e.g., \$50k) in a year.

 We study the certified robustness of three robust classifiers (RS \cite{cohen2019certified}, Local Lipschitz constrained \cite{huang2021training} and SPLITZ) on the Adult dataset shown in Fig \ref{fig:adult_cd}. We observe that SPLITZ achieves better tradeoffs between the certified robustness and accuracy.

\vspace{5pt}
\noindent\textbf{Results on Law School Dataset}
 Law School dataset includes the admission related data with 7 features (LSAT score, gender, undergraduate GPA etc.) of $N=4,862$ applicants ($N_{train} = 3 ,403$, $N_{test} = 1,459$) to predict the likelihood of passing the bar.

 Similar to the previous comparison, we evaluate SPLITZ against two other methods: Randomized Smoothing (RS) \cite{cohen2019certified} and Local Lipschitz Constraints \cite{huang2021training} as shown in Fig \ref{fig:law_cd}. Our findings demonstrate that SPLITZ consistently surpasses these techniques, achieving a significantly larger certified radius.}

\section{Discussion and conclusion}
In this paper, we presented \textit{SPLITZ}, a novel and practical certified defense mechanism, where we constrained the local Lipschitz bound of the left half of the classifier and smoothed the right half of the classifier with noise. This is because the local Lipschitz constant can capture  information specific to each individual input, and the relative stability of the model around that input. To the best of our knowledge, this is the first systematic framework to combine the ideas of local Lipschitz constants with randomized smoothing.
Furthermore, we provide a closed-form expression for the certified radius based on the local Lipschitz constant of the left half of the classifier and the randomized smoothing based radius of the right half of the classifier. We show that maintaining a relatively small local Lipschitz constant of the left half of the classifier helps to improve the certified robustness (radius).\revision{We showed results on several benchmark datasets and obtained significant improvements over state-of-the-art methods for the MNIST, CIFAR-10 and ImageNet datasets. For instance, on the CIFAR-10 dataset, \textit{SPLITZ} can achieve $43.2\%$ certified test accuracy compared to state-of-art certified test accuracy $39.8\%$ with $\ell_2$ norm perturbation budget of $\epsilon = 1$. We observed similar trends for the MNIST and ImageNet dataset.}  We believe that combining the core idea of \textit{SPLITZ} with other recent techniques, such as denoising diffusion models and adversarial training can be a fruitful next step to further improve certified robustness.

\begin{figure*}[t]
\centering
\includegraphics[scale=0.4]{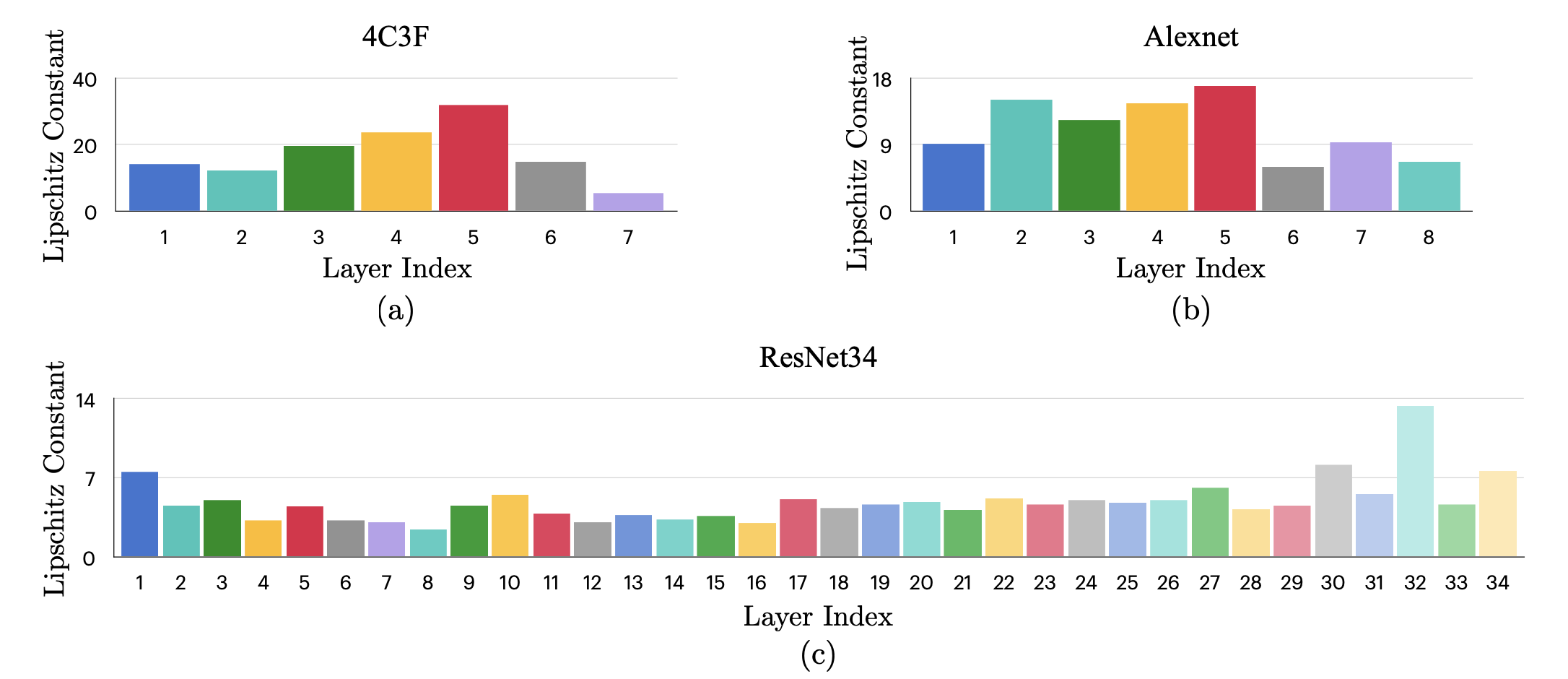}
\caption{Lipschitz Constants of each affine layer in pretrained models:(a) 4C3F model (there are 4 convolution layers and 3 fully- connected layers in the neural network. )  \cite{huang2021training} , (b) Alexnet model \cite{krizhevsky2012imagenet}, (c) ResNet34 model \cite{he2016deep}. We can observe the similar trends that right half of the model usually contain a larger Lipschitz constant, while the left half of the model preserves a relatively smaller Lipschitz constant.}
\label{fig: Lip const appen}
\vspace{-10pt}
\end{figure*}
\begin{figure*}[t]
    \centering
    \includegraphics[scale=0.4]{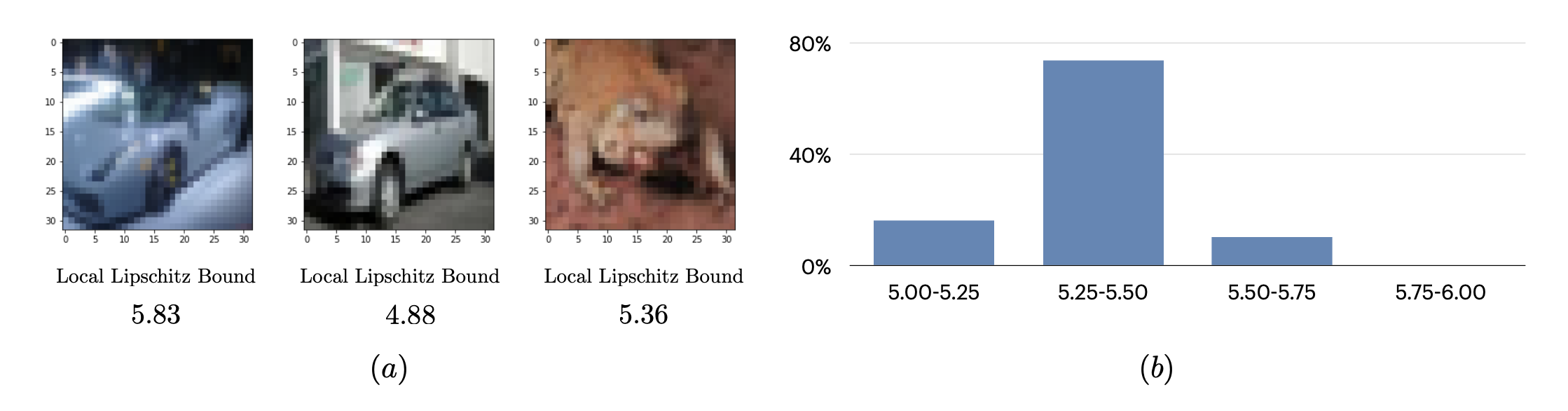}
    \caption{(a) Local Lipschitz bound for three random CIFAR-10 images on Alexnet, (b) Percentage analysis of local Lipschitz bound in CIFAR-10 test data.}
    \label{fig: app_lip_2}
\end{figure*}

\appendices
\vspace{-10pt}
\section{Additional Lipschitz Constants Results}\label{Sec: addtional_Lipschitz_results}

In this Section, we provide additional Lipschitz constants results in the prevalent neural networks in Fig \ref{fig: Lip const appen}. We can observe the similar trends as previous that the right half of the neural network is more \textit{unstable} than the right half of the neural network. As shown in Fig \ref{fig: app_lip_2} (a), we notice considerable variation in the values of local Lipschitz constants across different input images, a trend that is consistent throughout the entire CIFAR-10 test dataset as depicted in Fig \ref{fig: app_lip_2} (b). These findings lead us to reconsider the efficacy of directly smoothing the input. Such an approach doesn't cater to the observed heterogeneity. Alternatively, injecting noise at an intermediary step within the classifier can make the model more robust to disturbances.

\vspace{-10pt}
\section{Lipschitz Constrained Training} \label{sec:Lipschitz_bound_theory}
From the scope of this paper, we utilize the local Lipschitz constrained training for the left half of the classifier introduced in \cite{huang2021training}. We focus on $l_2$ norm denoted as $\parallel \cdot \parallel$. Now we consider a neural network $f$ containing $M$ affine layers (parameterized by $w$) each followed by a clipped version ReLU$\theta$, which is defined as follows:
\begin{align}
    ReLU\theta(x) = 
    \begin{cases}
        0, & \text{if}~~x \leq 0 \\
        x, & \text{if}~~0 < x < \theta \\
        \theta, & \text{if}~~ x \geq \theta \\
    \end{cases}
\end{align}
The neural network maps input $x$ to output $f(x)$ using the following architecture:
\begin{align}
    z_1 = x; z_m(x) = ReLU\theta (W_m x), z_{M+1} = W_M z_M
\end{align}
We define the perturbation around the input $x$ as:
\begin{align}
    x' = x + \epsilon, \parallel \epsilon \parallel \leq \delta, \delta\geq 0
\end{align}

By adding perturbation around input $x$ within a $\delta$ ball, $z(x')$ can be bounded element-wise as $LB \leq z(x') \leq UB$, where $LB$ and $UB$ can obtain by bound propagation methods \cite{gowal2018effectiveness,lee2020lipschitz}. We then define the diagonal matrix $I^v$ to represent the entries where ReLU$\theta$'s outputs are $varying$:
\begin{align}
    I^v(i,i) = 
    \begin{cases}
        1, & \text{if}~~UB_i > 0~~ and~~ LB_i < \theta\\
        0, & \text{otherwise}\\
    \end{cases}
\end{align}
Next, the output of the ReLU$\theta$ $D^v$ can be defined as follows:
\begin{align}
    D^v(i,i) = 
    \begin{cases}
        \mathbbm{1}(\text{ReLU}\theta(z_m^i)>0), & \text{if}~~I^v(i,i) =1  \\
        0, &\text{otherwise}\\
    \end{cases}
\end{align}
where $ \mathbbm{1}$ denote the indicator function. 
Then the local Lipschitz bound at input $x$ is:
\begin{align}
    L_{local}(x,f) \leq \parallel W_M I^v_{M-1} \parallel \parallel I^v_{M-1} W_{M-1} I^v_{M-2} \parallel \dots \parallel I^v_1 W_1 \parallel
\end{align}
As stated in \cite{huang2021training}, it straight forward to prove 
$\parallel I^v_{M-1} W_{M-1} I^v_{M-2}  \parallel \leq \parallel W_{M-1} \parallel$ using the property of eigenvalues. We briefly prove it following from \cite{huang2021training}.
\begin{proof}
    Let $W' = [W~I]^{T}$, The singular value of $W'$ is defined as the square roots of the eigenvalues of $W'^T~  W'$. We know the following
    \begin{align}
        {W'}^T W' = W^T W + I^T I \geq W^T W.
    \end{align}
    Therefore, we get the following result:
    \begin{align}
        \parallel W' \parallel \geq \parallel W \parallel
    \end{align}
    We complete the proof.
\end{proof}

Next, we will give a toy example to further illustrate the idea of local Lipschitz bound. 

\noindent\textbf{A toy example} Here we provide a similar toy example as mentioned in \cite{huang2021training}. Consider a $2$-layer neural network with ReLU$\theta$ activation layer:
\begin{align}
    x \rightarrow \text{Linear1}(W^1) \rightarrow \text{ReLU}\theta \rightarrow \text{Linear2}(W^2) \rightarrow y
\end{align}
where $x \in \mathcal{R}^3$ and $y \in \mathcal{R}$ and $W^m$ denotes the weight matrix for layer $m$. Moreover the threshold $\theta = 1$.

Given the input [1,-1,0] with $\ell_2$ perturbation 0.1. Assume the weight matrices are: 
\begin{align}
 W^1 = \begin{bmatrix}
2 & 0 & 0\\
0 & 2 & 0 \\
0 & 0 & 1
\end{bmatrix} , W^2 = [1,1,1]
\end{align}
Thus, we have the following:
\begin{align}
   &\text{Input}~[1,-1,0] \rightarrow \\ \nonumber&\begin{bmatrix}
[0.9 & 1.1 ]\\
[-1.1 & -0.9]   \\ 
[-0.1 & 0.1] 
\end{bmatrix} \times \begin{bmatrix}
2 & 0 & 0\\
0 & 2 & 0 \\
0 & 0 & 1
\end{bmatrix}
\rightarrow
\begin{bmatrix}
[1.8 & 2.2 ]\\
[-2.2 & -1.8]   \\ 
[-0.1 & 0.1] 
\end{bmatrix}
\end{align}
According to the above upper bound (UB) and lower bound (LB), we obtain the $I_V$ function as follows:
\begin{align}
    I_V^1 = \begin{bmatrix}
       0 & 0 & 0\\
        0 & 0 & 0 \\
        0 & 0 & 1 
    \end{bmatrix}
\end{align}
Overall, we have the local Lipschitz bound as follows:
\begin{align}
    L_{local}(x,f) \leq \parallel  W^2 I_V^1 \parallel\parallel I_V^1 W^1 \parallel = 1
\end{align}
For the global Lipschitz bound, we have the following:
\begin{align}
    L_{global} \leq \parallel  W^2 \parallel \parallel  W^1 \parallel = 4
\end{align}
Overall, we can find that the local Lipschiz bound is much tighter than the global Lipschitz bound.

\section{Additional Experimental Details}\label{sec:addtional experimental_results}
In this section, we provide additional results for the three datasets, e.g., MNIST, CIFAR-10 and ImageNet dataset. We first provide the details of three datasets. Next, we illustrate the baselines used in our paper. Note that we report the numbers (certified test accuracy, average certified radius) from respective papers. 

\noindent\textbf{Training details}
\revision{For all value of $\sigma$, we keep the value of training $\sigma$ and testing $\sigma$ to be the same. We apply the noise samples $n_{0} = 100$ to predict the most probably class $c_A$ and denote $\alpha=0.001$ as the confidence during the certifying process. Furthermore, we use $n_{1} = 100000, 100000, 10000$ to calculate the lower bound of the probability $p_A$ for the MNIST, CIFAR-10 and ImageNet dataset respectively. Moreover, to maintain a relatively small local Lipschitz constant of left half of the \textit{SPLITZ} classifier, we set the threshold of clipped ReLU (see Sec \ref{sec:Lipschitz_bound_theory}) as 1 for all datasets. For estimating the local Lipschitz constant of the left half of the classifier, the power iteration is 5, 5, 2 during the training for MNIST, CIFAR-10 and ImageNet respectively following from \cite{huang2021training}.}

 \begin{table}[t]
    \centering
    \begin{tabular}{l c c c c c c c}
        \hline
         MNIST& & \multicolumn{5}{c}{Certified Test Accuracy at $\epsilon~(\%)$}\\
         \cline{3-7}
         Method &  Extra data & 1.50 & 1.75 & 2.00 & 2.25 & 2.50   \\
         \hline
         \hline
         RS \cite{cohen2019certified}& \xmark & 67.3 & 46.2 & 32.5 & 19.7 & 10.9 \\ 
         MACER \cite{Zhai2020MACER:}& \xmark & 73.0 & 50.0 & 36.0 & 28.0 & - \\
         Consistency \cite{jeong2020consistency}& \xmark & 82.2 & 70.5 & 45.5 & 37.2 & 28.0 \\
         SmoothMix \cite{jeong2021smoothmix}& \xmark  & 81.8 & 70.7 & 44.9 & 37.1 & 29.3 \\
         DRT \cite{yang2021certified} & \xmark  & 83.3 & 69.6 & 48.3 & 40.3 & 34.8\\
         RS+OrthoNN \cite{wang2020orthogonal}  & \xmark & 70.1 & 49.7 & 33.2 & 21.0 & 10.5 \\
         \hline
         \textit{\textbf{SPLITZ}} & \xmark & 80.2 & \textbf{71.3} & \textbf{62.3} & \textbf{51.7} & \textbf{40.5} \\
         \hline
    \end{tabular}
    \caption{\revisionfinal{Comparison of certified test accuracy (\%) on MNIST under $\ell_2$ norm perturbation. Each entry lists the certified accuracy using numbers taken from respective papers (RS results follow from previous benchmark papers \cite{jeong2020consistency,jeong2021smoothmix}).}}
    \label{tab:mnist results_2}
\end{table}
\noindent\textbf{Results on MNIST} 
\revisionfinal{As showed in Table \ref{tab:mnist results_2}, we can observe that \textit{SPLITZ} outperforms other state-of-art approaches in almost every value of $\epsilon$. Impressively, we find that the \textit{SPLITZ} classifier has a significant improvement when the value of $\epsilon$ is large. For instance, when $\epsilon = 2.50$, \textit{SPLITZ} classifier achieves \textbf{40.5\%} compared to
state-of-art top-1 test accuracy 34.8\% certified test accuracy on the MNIST dataset. Moreover, when we increase $\epsilon$ from 1.50 to 2.50, RS drops from 67.3\% to 10.9\% decreasing 56.4\% test accuracy. \textit{SPLITZ}, however, maintains higher certified test accuracy from 80.2\% to 40.5\% maintaining relatively higher test accuracy.}

\bibliographystyle{ieeetr}
\bibliography{reference}
\vspace{-20pt}
\begin{IEEEbiography}[{\includegraphics[width=1in,height=1.25in,clip,keepaspectratio]{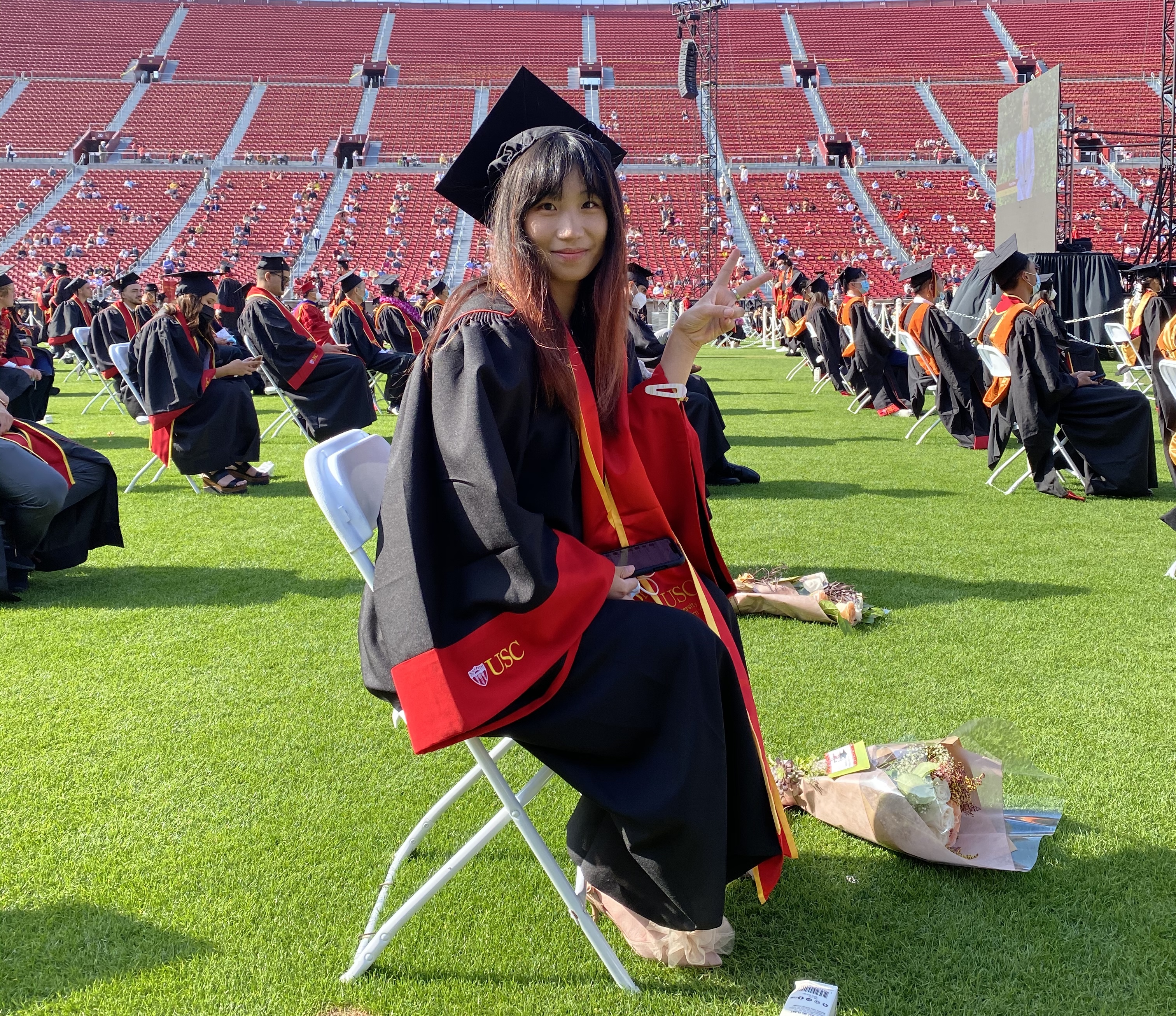}}]{Meiyu Zhong}
received the bachelor's from Shanghai University, China in 2019 and her master's degree from University of Southern California, CA, USA, in 2021. She started the Ph.D. program Electrical and Computer Engineering from The University of Arizona in 2021. She has interned at Micron Technology in 2024 and 2025. Her current research interests include trustworthy machine learning with a particular focus on robustness and fairness in machine learning, as well as large language model efficiency.
\end{IEEEbiography}
\vspace{-20pt}
\begin{IEEEbiography}[{\includegraphics[width=1in,height=1.25in,clip,keepaspectratio]{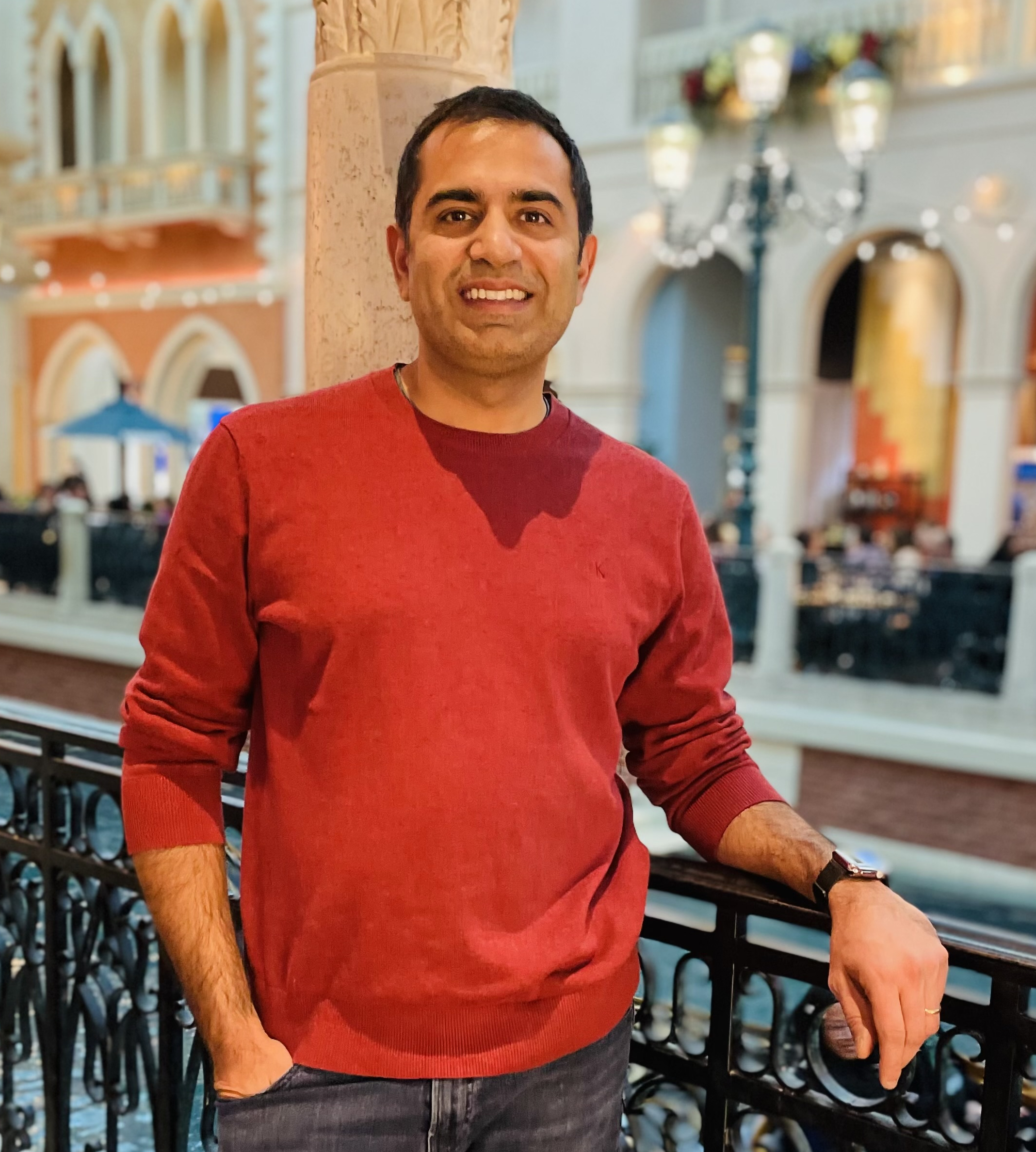}}]{Ravi Tandon} (Senior Member, IEEE)
received the B.Tech. degree in electrical engineering from Indian Institute of Technology, Kanpur (IIT Kanpur), in 2004, and the Ph.D. degree in electrical and computer engineering (ECE) from the University of Maryland, College Park (UMCP), in 2010. He is currently a Professor in the Department of ECE at the University of Arizona. He has also held positions with the Bradley Department of ECE, Hume Center for National Security and Technology; and the Department of Computer Science, Discovery Analytics Center at Virginia Tech. From 2010 to 2012, he was a Post-Doctoral Research Associate with Princeton University. His current research interests include information theory and its applications to wireless networks, signal processing, communications, security and privacy, machine learning, and data mining. He was a recipient of the 2018 Keysight Early Career Professor Award, an NSF CAREER Award in 2017, and the Best Paper Award at IEEE GLOBECOM 2011. He has served on the Editorial Board of IEEE TRANSACTIONS ON WIRELESS COMMUNICATIONS and is currently an Editor of IEEE TRANSACTIONS ON COMMUNICATIONS and IEEE TRANSACTIONS ON INFORMATION THEORY.
\end{IEEEbiography}
\end{document}